%% file: FedALC_new.tex
\documentclass[journal]{IEEEtran}
\usepackage{amsmath,amsfonts}
\usepackage{algorithmic}
\usepackage{algorithm}
\usepackage{array}
\usepackage[caption=false,font=normalsize,labelfont=sf,textfont=sf]{subfig}
\usepackage{textcomp}
\usepackage{stfloats}
\usepackage{subfig}
\usepackage{subfloat}
\usepackage{url}
\usepackage{verbatim}
\usepackage{graphicx}
\usepackage{cite}
\usepackage{booktabs} 
\usepackage{amsfonts}
\usepackage[utf8]{inputenc}

\usepackage{amsmath}
\usepackage{stmaryrd}

\usepackage{amssymb}
\usepackage{verbatim}
\usepackage{multirow}

\usepackage{color}
\usepackage{bm}
\newtheorem{definition}{Definition}

\begin{document}

\title{Federated Learning with Only Positive Labels by Exploring Label Correlations}
\author{Xuming~An,
        Dui~Wang,
        Li~Shen,
        Yong~Luo,
        Han~Hu,
        Bo~Du,
        Yonggang~Wen~\IEEEmembership{Fellow,~IEEE},
        Dacheng~Tao~\IEEEmembership{Fellow,~IEEE}
\thanks{X. An and H. Hu are with the School of Information and Electronics, Beijing Institute of Technology, Beijing 100081, China (email: anxuming@bit.edu.cn; hhu@bit.edu.cn).}
\thanks{D. Wang, Y. Luo and B. Du are with the National Engineering Research Center for Multimedia Software, School of Computer Science, Institute of Artificial Intelligence and Hubei Key Laboratory of Multimedia and Network Communication Engineering, Wuhan University, Wuhan 430072, China (e-mail: wangdui@whu.edu.cn; yluo180@gmail.com; dubo@whu.edu.cn).}
\thanks{L. Shen and D. Tao are with the JD Explore Academy, Beijing, 100000, China (email: mathshenli@gmail.com, dacheng.tao@gmail.com).}
\thanks{Y. Wen is with the School of Computer Science and Engineering, Nanyang Technological University, Singapore 639798 (e-mail: ygwen@ntu.edu.sg).}
}
\markboth{$>$ \normalsize {TNNLS-2022-P-21572.R1} $<$}
{Shell \MakeLowercase{\textit{et al.}}: Bare Demo of IEEEtran.cls for IEEE Journals}




\maketitle

\input{text/abstract}

\input{text/introduction}
\input{text/related_work}

\input{text/problem_setup}

\input{text/methodology}

\input{text/experiments}

\input{text/conclusion}

\bibliographystyle{IEEEtran}
\bibliography{FedALC_new}

\end{document}

%% file: text/abstract.tex
\begin{abstract}

Federated learning aims to collaboratively learn a model by using the data from multiple users under privacy constraints. In this paper, we study the multi-label classification problem under the federated learning setting, where trivial solution and extremely poor performance may be obtained, especially when only positive data w.r.t. a single class label are provided for each client. This issue can be addressed by adding a specially designed regularizer on the server-side. Although effective sometimes, the label correlations are simply ignored and thus sub-optimal performance may be obtained. Besides, it is expensive and unsafe to exchange user's private embeddings between server and clients frequently, especially when training model in the contrastive way. To remedy these drawbacks, we propose a novel and generic method termed Federated Averaging by exploring Label Correlations (FedALC). Specifically, FedALC estimates the label correlations in the class embedding learning for different label pairs and utilizes it to improve the model training. To further improve the safety and also reduce the communication overhead, we propose a variant to learn fixed class embedding for each client, so that the server and clients only need to exchange class embeddings once. Extensive experiments on multiple popular datasets demonstrate that our FedALC can significantly outperform existing counterparts.

\end{abstract}

%% file: text/introduction.tex
\section{Introduction}
\label{sec:Introduction}

Federated learning (FL)~\cite{mcmahan2017communication} is a novel machine learning paradigm that trains an algorithm across multiple decentralized clients (such as edge devices) or servers without exchanging local data samples. Since clients can only access the local datasets, the user's privacy can be well protected, and this paradigm has attracted increasing attention in recent years~\cite{yang2019federated, li2020federated,li2021survey}.
In this paper, we study the challenge problem of learning a multi-label classification model~\cite{liu2017deep,shi2019mlne} under the federated learning setting, where each user has only local positive data related to a single class label~\cite{yu2020federated}. This setting can be treated as the extremely label-skew case in the data heterogeneity of federated learning, which is popular in real-world applications. For example, a user verification (UV) model is decentralized and cooperatively trained by aggregating sensitive information of multiple users, and each client has only access to specific information (such as voice~\cite{2017Deep}, face~\cite{2018Additive}, fingerprint\cite{Cao2019Automated} and iris\cite{nguyen2017iris}) of a single user.

The key challenges for federated learning with only positive labels is that collapsing may occur for embedding-based discriminative models~\cite{vaswani2017attention, konevcny2016federated}, such as the popular neural network based classifier.
In our setting, each client lacks negative information including negative instances and negative class embeddings. Naively employing vanilla federated learning algorithms such as Federated Averaging (FedAvg), may lead to the collapsing of class embeddings and result in a trivial solution or extremely poor performance~\cite{bojanowski2017unsupervised}. This is because that the client neglects the negative loss and only indiscriminately minimizes the distance between the positive instance embedding and class embedding. In this case, different class scores on a test sample will be similar and indistinguishable, and hence the resulting model is meaningless.

This issue is investigated and initially addressed by a recent work \cite{yu2020federated}, where an algorithm termed Federated Averaging with Spreadout (FedAwS) is proposed by imposing a geometric regularization on the server-side to separate different class embeddings and prevent them collapsing to a single point.
This approach, however, treats different labels equally in the spreadout (class embedding separation) process. That is, embeddings of class labels that are highly correlated and significantly different in multiple labels' space are separated in the same way. This is not reasonable since embeddings should be close for correlated labels, and dissimilar otherwise. For example, we assume that the class labels `Desktop computer' and `Desk' often appear in the same instance, thus these two corresponding class embedding vectors can be deemed high-correlation and may be relatively close compared with others, such as class labels `aircraft', `automobile', etc. Besides, since the instance and class embeddings are trained on clients and server respectively, the class embeddings should be transmitted frequently between the server and clients. This leads to not only high communication and training time costs, but also privacy leakage in the transmission process. The leakage risk would increase significantly with an increasing number of transmissions.

\begin{figure*}[t]   
    \begin{center}      
    \includegraphics[width=2\columnwidth]{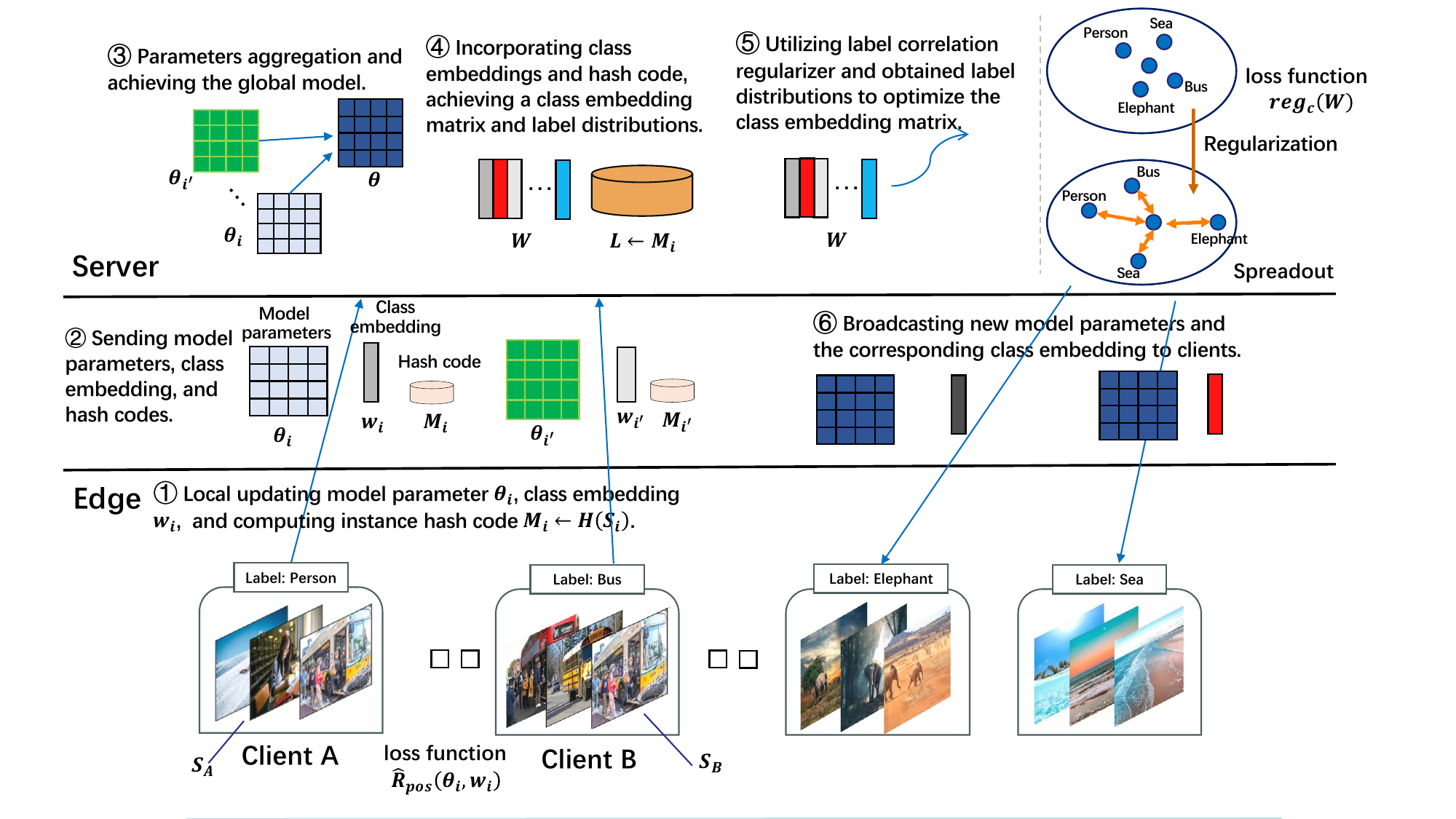}    
    \end{center}
    \caption{{\textbf{Overview of the proposed federated averaging by exploring label correlations (FedALC) method}. \textbf{(1)} Our FedALC computes gradients for parameter updating and hash code for each instance locally. The instance hash code is utilized for calculating label correlations and \textbf{only need transmission once}; \textbf{(2)} The client sends the locally updated model parameters, class embedding, and hash codes to the server; On the server, \textbf{(3)} the global model is obtained via parameter aggregation; \textbf{(4)} the different class embeddings are merged as a matrix, and label distribution is obtained by comparing the hash codes; \textbf{(5)} The server then utilizes our designed correlation regularizer based on the label distribution to optimize the class embedding matrix; \textbf{(6)} and eventually transmits global model parameters and corresponding class embeddings to different clients.}}
    \label{pipeline}
\end{figure*}

Given the issue of model collapsing and limitations of prior art, we propose a novel method termed Federated Averaging by exploring Label Correlations, named FedALC (see Fig.~\ref{pipeline}), for FL with only positive labels. In particular, we construct a novel regularizer to optimize the class embedding matrix by taking the label correlations into consideration. This is achieved by adding a correlation-based weight for each label pair in the spreadout, so that embeddings of similar labels are relatively close and dissimilar labels are far away. In this way, the label correlation can be well considered, and meanwhile the issue of collapsing can be addressed. To obtain the label correlations for regularization, we also propose an encrypted transmission strategy to collect label information across clients and obtain label distributions on the server. The label distributions are then utilized for calculating the weight of the correlation between different labels. Besides, to reduce both the computational cost and transmission overhead, we propose a variant to learn fixed class embedding matrix so that only a single transmission of class embeddings between the server and clients should be conducted.

To summarize, the main contributions of this paper are:
\begin{itemize}
  \item We propose a novel method termed FedALC to exploit the correlations between different class labels for federated learning with only positive labels. Our method can make the embeddings of relevant labels close to each other and stay away from irrelevant labels in class embedding learning;
  \item We design an encrypted and communication-efficiency strategy for calculating label correlations by collecting labels across clients and constructing label distributions on the server;
  \item We propose a variant that learns fixed class embeddings to improve safety and reduce transmission overhead and training cost.
\end{itemize}
We conduct extensive experiments on three popular visual datasets and two challenging multi-label text datasets. The results demonstrate significant improvements over the competitive counterparts. For example, we achieve relative improvements of $8.17\%$ and $19.3\%$ on the VOC 2012~\cite{everingham2010pascal} and Bibtex~\cite{2008Multilabel} datasets respectively compared with FedAwS.

The rest of paper is organized as follows. In section~\ref{sec:Related_Work}, we review related works about multi-label learning and federated learning. Section~\ref{sec:Problem_Formulation} introduces the problem of multi-label federated learning with positive labels. Details of the proposed FedACL method is presented in Section~\ref{sec:Method_FedALC}. The experimental results and analysis are shown in ~\ref{sec:Experiments}, and we conclude this paper in~\ref{sec:Conclusion}.

%% file: text/related_work.tex
\section{Related Work}
\label{sec:Related_Work}

\subsection{Multi-label learning}
Multi-class learning aims to classify instances into one of multiple classes. In contrast to the traditional multi-class classification, where only a single class label is assigned to an instance~\cite{wang2018baseline,tsoumakas2009mining,zhang2013review}, a single instance may have more than one label in multi-label classification (MLC), which can be utilized in a variety of applications ranging from protein function classification~\cite{barutcuoglu2006hierarchical}, document categorization~\cite{yang2016hierarchical}, to automatic image annotation~\cite{wang2016cnn,9721413}. A dozen of approaches have been proposed for various settings of MLC. 
For example, partial multi-label~(PML)~\cite{xie2018partial,xie2021partial} studies a problem that a superset of labels is given, where some irrelevant labels may be contained in the superset. Multi-label learning with weak labels~\cite{sun2010multi} assumes that only a subset of labels is provided. That is, some proper labels are absence for an instance. A thorough review of multi-label methods can be found in~\cite{liu2021emerging}.
Besides, in this era of big data, the extreme multi-label learning~(XMLC)~\cite{jain2016extreme,9758948} that handles a large number of labels has attracted more and more attention and becomes a new mainstream direction of MLC.

\subsection{Federated learning}
Federated learning~(FL) is a distributed machine learning paradigm that multiple clients collaborate to complete a learning task without exchanging private data, while maintaining communication efficiency. A well-known and commonly used FL algorithm is Federated Averaging~(FedAvg)~\cite{mcmahan2017communication}, which learns a global model by simple averaging.
Recently, numerous and more sophisticated algorithms or aggregation strategies have been proposed to address various FL challenges. For example, the performance of local models may be better than the global model due to the distribution divergence issue \cite{xu2021vitae,zhang2022vitaev2}, and personalized federated learning~\cite{2021Exploiting,2021Personalized,2020Ditto,9743558} was proposed to address this issue by adopting a personalized method to optimize the global model for each client.
In~\cite{2012Privacy,mcmahan2017communication,2019Federated}, some new approaches are designed for better privacy protection, while other works \cite{2020Acceleration,2020Qsparse,xu2020ternary} mainly focused on reducing communication cost. More recent works assume a heterogeneous setup that data on different clients hold different distributions~\cite{2021Data,2021Bias,2020KD3A,sattler2019robust}. Client clustering and selection are also studied~\cite{2021Heterogeneity,2021Clustered,jamali2022federated,li2022data}. In this paper, we study a challenging FL setting that each client has only access to positive data~\cite{yu2020federated}.

The most related work to our proposed FedALC is FedAwS \cite{yu2020federated}. However, our work is different from the former work in the following aspects: 1) FedAwS simply constructs a regularizer to make different classes separate, while our method constructs the regularizer in a more fine-grained manner by considering the label correlations; 2) To construct our correlation regularizer on the server, a novel strategy is designed to collect label correlations across clients with the guarantee of safety and communication efficiency; 3) To further improve safety and reduce communication overhead, we also develop a variant of our approach to learn fixed class embedding for each client, and thus client and server only need to transmit class embedding once.

%% file: text/problem_setup.tex
\section{Problem Formulation}
\label{sec:Problem_Formulation}

We consider the embedding-based model~\cite{vaswani2017attention, konevcny2016federated} to complete multi-label classification task under the federated learning setting, where each client has only access to the local dataset associated with a single label. We suppose that there are $m$ clients, and the $i$-th client is associated with a single label $y^{i}$. In multi-label classification, an instance may have multiple labels, and hence the same instance may be distributed in different clients. For $i \in [m]$, the ${i}$-th client has access to the local dataset $\mathcal{S}^{i}=\{(\mathbf{x}_{1}^{i},y^{i}),(\mathbf{x}_{2}^{i},y^{i}),\cdots,(\mathbf{x}_{n_{i}}^{i},y^{i})\} $. An embedding network $g_{\theta}: \chi \rightarrow \mathbb{R}^D$ maps the input $\mathbf{x}$ as the $D$-dimensional instance embedding $g_{\theta}(\mathbf{x})$. The label of all $n_{i}$ instances are the same since each client only has a single label.
Given a distance measure $d(\cdot,\cdot)$, the contrastive loss~\cite{oord2018representation} used in clients can be given by:
\begin{equation}
\begin{split}
\ell_{\mathrm{cl}}(\mathbf{x}_j^{i},y^{i}; \theta,W)=\alpha \cdot\left(d\left(g_{\boldsymbol{\theta}}(\mathbf{x}_j^{i}), \mathbf{w}_{y^{i}}\right)\right)^{2} +\quad \quad \quad\quad \quad\\
\quad \quad \quad \beta \cdot \sum_{c \neq i}\left(\max \left\{0, \nu-d\left(g_{\boldsymbol{\theta}}(\mathbf{x}_j^{i}), \mathbf{w}_{y^{c}}\right)\right\}\right)^{2}\quad,
\end{split}
\label{client_loss_expected}
\end{equation}
where $\alpha$ and $\beta$ are trade-off hyper-parameters, $\theta$ is parameters of model (feature extractor), $W \in \mathbb{R}^{C \times D}$ is the class embedding matrix ($C=m$ in our setting) and $D$ is the dimension of embedding, with each row being a class embedding $\mathbf{w}_{y^{c}}$. The first and second terms of Eq.~(\ref{client_loss_expected}) are the positive loss and negative loss, respectively;
{$v$ is the edge hyper-parameter, which is a margin for the maximum distance between the positive label and the negative label embedding to stabilize model training\cite{oord2018representation}.}
Then, the prediction function can be expressed as:
\begin{equation}
    f(\mathbf{x})=W g_{\boldsymbol{\theta}}(\mathbf{x}).
\end{equation}
{The output of $f(x)$ yields a real valued tensor, with each slot corresponding to a specific class. Following a methodology similar to FedAwS algorithm, we employ the top-k largest value method to predict the labels.}
Since the client has no access to the instances of negative classes in our setting, we rewrite the loss function Eq.~(\ref{client_loss_expected}) as the following form by dropping the negative loss:
\begin{equation}
{\ell^{\mathrm{pos}}_{cl}(\mathbf{x}_j^{i}, y^{i})=\left(d\left(g_{\boldsymbol{\theta}}(\mathbf{x}_j^{i}), \mathbf{w}_{y^{i}}\right)\right)^2.}
\label{client_loss}
\end{equation}
Then the empirical loss on the local dataset $\mathcal{S}^{i}$ can be given by:
\begin{equation}
{\hat{\mathcal{R}}_{\mathrm{pos}}\left(\mathcal{S}^{i}, y^{i}\right):=\frac{1}{n_{i}} \sum_{j \in\left[n_{i}\right]} \ell^{\mathrm{pos}}_{cl}\left(\mathbf{x}_j^{i},y^{i}\right).}
\end{equation}

To avoid the collapsing of class embedding matrix in our setting, we adopt the classic strategy that adds a geometric regularizer to let the matrix be spreadout, and the whole objective function is given as follows:
\begin{equation}
\begin{split}
&\qquad\qquad\qquad \quad \min \limits_{\theta, W} \mathcal{L}(\mathcal{S}; \theta, W),\\
&\operatorname{s.t.} {\ \mathcal{L}}:= \sum_{i \in[m]}\frac{|\mathcal{S}^{i}|}{|\mathcal{S}|} \hat{\mathcal{R}}_{\mathrm{pos}}\left(\mathcal{S}^{i}, y_{i}\right)+\lambda \cdot \mathrm{reg}(W),\\
\end{split}
\end{equation}
where $\mathcal{S}$ is {the union} of all $\mathcal{S}^{i}$, and $\lambda \geq 0$ is the balancing parameter. The objective is to minimize both the sum of all clients' empirical risks and the structure risk w.r.t. the class embedding matrix. The former term encourages instance embeddings $g_{\boldsymbol{\theta}}(\mathbf{x}^{i})$ to be close to the corresponding class embedding $\mathbf{w}_{y^{i}}$, which is performed on the client. The latter disperses the class embedding matrix, which is performed on the server. 


%% file: text/methodology.tex
\section{Federated Averaging by Exploring Label Correlations}
\label{sec:Method_FedALC}

{In this section, we propose methods that leverage label correlations to alleviate the issue of the model collapsing in FL with only positive labels. Initially, we introduce the most related work FedAwS, which employs a geometric regularizer to encourage class embeddings to be distinct from each other. Subsequently, we present the Federated Averaging by exploring the Label Correlations (FedALC) algorithm, wherein the server optimizes a correlation regularizer utilizing negative label pairs in each FL round. Through the correlation regularizer, the server can obtain a matrix where positive class embeddings are distanced
from negative class embeddings. To expedite convergence and reduce the communication round, we devise a novel correlation regularizer that learns a fixed class embedding matrix by incorporating both positive and negative label pairs before the model training (FedALC-fixed). We then conduct an analysis of the convergence and optimality of our proposed method. Lastly, we introduce a label collection method to obtain the label correlations safely.}

\subsection{Discussion on FedAwS}
Based on the FedAvg framework~\cite{mcmahan2017communication}, a recent work termed Federated Averaging with Spreadout (FedAwS)~\cite{yu2020federated} is presented to handle this federated setting that clients have only access to the positive data for a single class. To address the challenge, the FedAwS framework uses a spreadout strategy to optimize the class embedding matrix, which can disperse the embedding matrix and alleviate the issue of collapsing. This is achieved by letting the server perform an geometric regularizer to ensure that class embeddings are separated from each other, in addition to averaging shared model parameters.

This process can be formulated as applying the following geometric regularization in each communication round on the server:
\begin{equation}
\operatorname{reg}_{\mathrm{sp}}(W)=\sum_{u \in[K]} \sum_{u^{\prime} \neq u}\left(\max \left(0, \nu-d\left(\mathbf{w}_{u}, \mathbf{w}_{u^{\prime}}\right)\right)\right)^{2},
\end{equation}
where $K$ is the number of classes. Since the hyper-parameter $\nu$ may be very hard to choose, FedAwS also reformulate (3) as:
\begin{equation}
\operatorname{reg}_{\mathrm{sp}}^{\mathrm{top}}(W)=\sum_{u\in [K]}\sum_{u' \in \mathcal{K}', \atop u' \neq u}-\left(d\left(\mathbf{w}_{u}, \mathbf{w}_{u'}\right)\right)^{2} \cdot \llbracket u' \in \mathcal{N}_{k}(u) \rrbracket,
\end{equation}
where $\mathcal{K}'$ is a subset of classes, and $\mathcal{N}_{k}(u)$ is the set of $k$ classes that are close to the class $u$ in the embedding space. 
{The relationship of equation (11) and (12) comes from \cite{yu2020federated}. According to Section 4.2 in \cite{yu2020federated}, by adaptively setting hyperparameter $v$ to be the distance between the class $c$'s embedding $w_c$ and its $(k+1)-$th closest class's embedding, we have the term $\max\left\{0, v-d(w_c, w_{c^{'}})\right\} = 0, \forall c^{'} \notin \mathcal{N}_k(c)$ and $\max\left\{0, v-d(w_c, w_{c^{'}})\right\} = v-d(w_c, w_{c^{'}}), \forall c^{'} \in \mathcal{N}_k(c)$. Since $v$ is a constant, minimization of $v-d(w_c, w_{c^{'}})$ is consistent with minimizing $-d(w_c, w_{c^{'}})$, (11) is equivalent with (12).}
This is motivated by the stochastic negative mining approach~\cite{reddi2019stochastic}. Since only the distances between close class embeddings are maximized, the computational cost can be reduced. 

A main drawback of FedAwS is that the embeddings of different classes are separated in the same way without considering the correlations between different class labels as mentioned above. The simple spreadout regularizer is effective in the task of multi-class classification but may be coarse in multi-label task, where some labels are usually highly correlated, especially in the XMLC~\cite{jain2016extreme} scenario. This issue is appropriately tackled by exploring label correlations in our method. {In comparison to FedAwS, the proposed FedALC (or FedALC-fixed) holds the advantage of leveraging label correlation to optimize the class embedding matrix. The assessment of correlation among multiple labels is accomplished by tallying the frequency of the labels cooccuring with the same instance. To safeguard against the server discerning the specific labels tagged to a given instance, in addition to hashing the instance, the labels themselves can also be hashed. Further elaboration on this method is provided in Section \ref{collect}. This approach ensures that the server can ascertain label correlations while remaining oblivious to the association between instances and specific labels.}

\subsection{Spreadout by exploring label correlations (FedALC)}
\begin{figure}[t]
    \begin{center}
    \includegraphics[width=1.0\columnwidth]{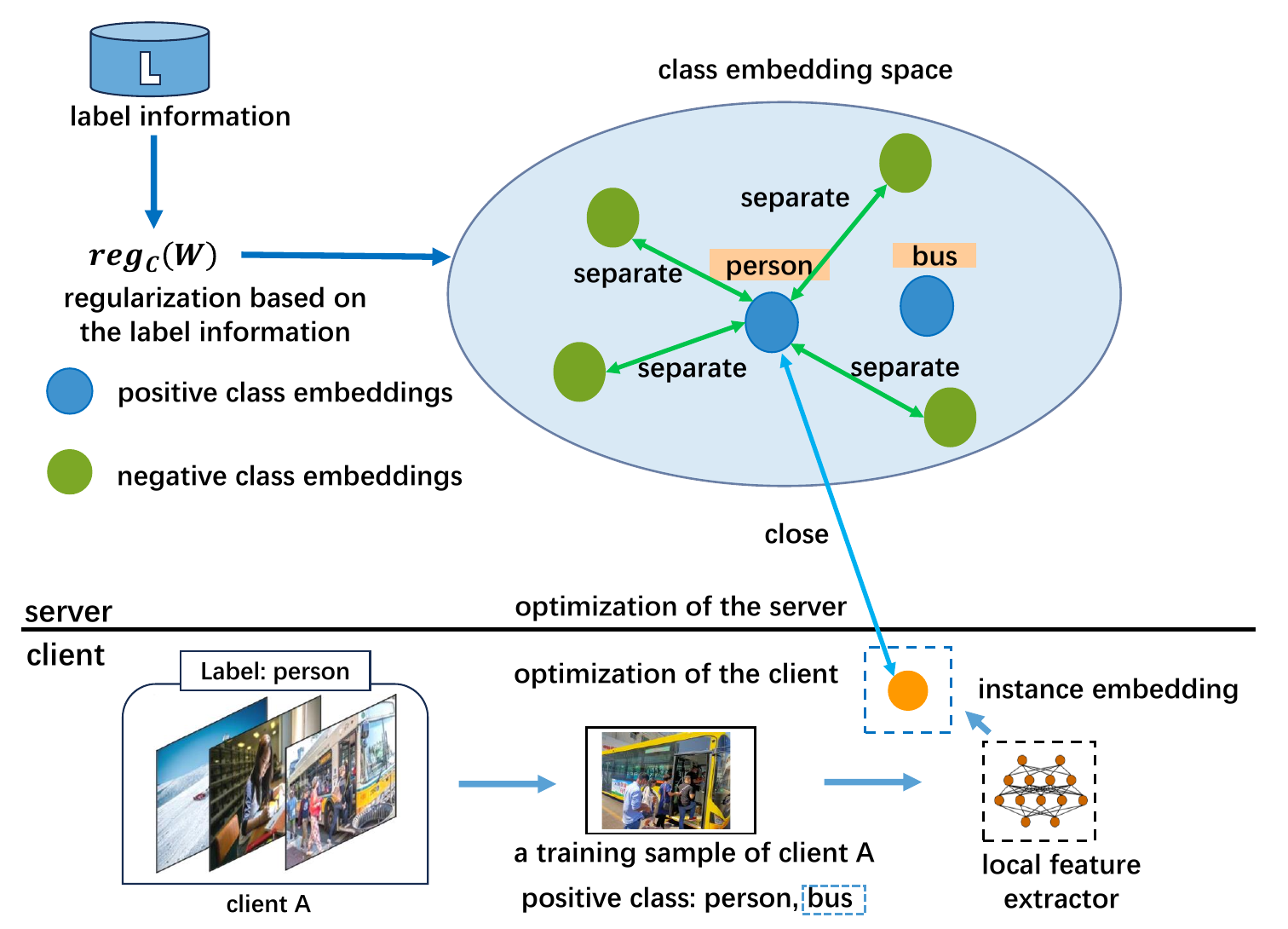}
    \end{center}
    \caption{{Mechanism of the designed correlation regularizer under the federated learning paradigm. The orange, blue and green dots indicate the instance, positive class and negative class respectively.}}
    \label{fig:Norm_Fed_Contrast}
\end{figure}

{The correlation of class(label) has two meanings: Firstly, if two labels consecutively arise in one sample, then the correlation of the two labels are strong; alternatively, if they appear infrequently together, then the correlation is weak. Secondly, there are semantical connections between the classes (labels) such as, similarity and hierarchical relationship. For instance, the labels "train" and "car" as well as "train" and "vehicle" show strong correlations, whereas the labels "train" and "elephant" have a weak correlation.}

To exploit label correlations, we need to first collect the label set for each instance. This process will be described later. Given the collected label sets, we propose to add a correlation regularizer on the server-side to improve the dispersion of class embedding matrix by taking the label correlations into consideration. We first define $L_{j}^{P}$ and $L_{j}^{N}$ as the positive and negative label set for the $j$-th instance, respectively. For example, in a five-class classification problem, there is an instance with labels $\{0,1,2\}$, and then its label set is $L_{j}=[L_j^{P},L_j^{N}]$, where the positive label set $L_{j}^{P}=[0,1,2]$ and negative label set $L_{j}^{N}=[3,4]$. We collect the label sets of all clients' instances and merge them on the server, and the resulting whole label set is: ${L}=\{(L_{1}^{P},L_{1}^{N}),(L_{2}^{P},L_{2}^{N}),...,(L_{n_\mathrm{all}}^{P},L_{n_\mathrm{all}}^{N})\}$, where $n_\mathrm{all}$ is number of all clients' instances. Then the correlation regularizer can be given as follow:
\begin{equation}
\label{eq:corr_reg}
\begin{split}
&\operatorname{reg}_{\mathrm{cr}}(W)
=\sum_{i \in[m]}\frac{|\mathcal{S}^{i}|}{|\mathcal{S}|} \hat{\mathcal{R}}_{\mathrm{neg}}\left(\mathcal{S}^{i}\right)\\
&\!=\!\!  \frac{1}{n_{all}}\sum\limits_{i\in [m]}\sum\limits_{j\in [n_{i}]}\ell^{\mathrm{neg}}_{cl}({L_{j}^{P}}, {L_{j}^{N}},W)\!\!\\
&\!=\!\! \frac{1}{n_{all}} \sum\limits_{j\in [n_\mathrm{all}]}\!\! \left( \sum\limits_{y_j \in L_{j}^{P},\atop y_j^{\prime} \in L_{j}^{N}}\! \left(\max \left\{0, \nu \!-\!d(\mathbf{w}_{y_j}, \mathbf{w}_{y_j^{\prime}})\right\}\right)^{2}\right)\\
&\!=\!\!\sum_{u \in[C]}\!\! \left(\sigma_{uu'}\cdot \sum_{u^{\prime} \neq u}
\left(\max \left(0, \nu\!-\!d(\mathbf{w}_{u}, \mathbf{w}_{u^{\prime}})\right)\right)^{2} \right),
\end{split}
\end{equation}
where $\sigma_{uu'}$ is a weight that reflects the correlation between different class labels, and can be calculated as:
\begin{equation}
\sigma_{u u^\prime}=\frac{1}{n_{all}}\sum_{j\in [n_\mathrm{all}]}I\left((u \in L_j^P) \And (u' \in L_j^N)\right),
\end{equation}
where $I(\cdot)$ is an indicator function ($I(E)=1$ if event $E$ is true, and $0$ otherwise). The weight of label correlations is calculated based on the distribute of labels over the dataset. The weight is larger for more dissimilar labels and thus the dispersion would focus more on these labels. Since the number of instances for different labels may vary significantly, we add a normalization term to balance different labels, i.e.,
\begin{equation}
\begin{split}
\operatorname{reg}_{\mathrm{cr}}(W)
&\!=\!\!\sum_{u \in[C]}\!\left(\gamma_{uu'} \sum_{u^{\prime} \neq u}\left(\max\{0, \nu\!-\!d\left(\mathbf{w}_{u}, \mathbf{w}_{u^{\prime}}\right)\}\right)^{2} \right),
\end{split}
\end{equation}
where $\gamma_{uu'} = \frac{\sigma_{uu'}}{\sum_{u^{\prime} \neq u} {\sigma_{uu'}}}$.
To avoid the exhaustive tuning of hyper-parameter $\nu$, we follow~\cite{reddi2019stochastic, yu2020federated} to employ a negative mining strategy and reformulate (\ref{eq:corr_reg}) as:
\begin{equation}
\begin{split}
\operatorname{reg}_{\mathrm{cr}}^{\mathrm{top}}(W)
&\!=\!\!\sum_{u \in[C]}\!\! \left(\sigma_{uu'} \sum_{u^{\prime} \neq u,\atop u^{\prime} \in \mathcal{N}_{\mathcal{K}}(u) }
\left(\max \left(0, \nu\!-\!d(\mathbf{w}_{u}, \mathbf{w}_{u^{\prime}})\right)\right)^{2} \right),
\end{split}
\end{equation}
where $\mathcal{K}$ is the number of a subset of classes, and $\mathcal{N}_{k}(u)$ is the set of $\mathcal{K}$ classes that are most close to the class $u$ in the embedding space. We can also normalize the weight of correlations $\sigma_{uu'}$ to obtain $\gamma_{uu'}$ in this case. From the instance viewpoint, our correlation regularizer regards the set of positive class embeddings as the surrogate of its instance embeddings in the client negative loss of Eq.~(\ref{client_loss_expected}). Hence, the surrogate negative loss doesn't need instance embeddings and only needs class embeddings, which can be accessed in the server. According to the class embedding matrix and collected label sets, the surrogate negative losses of all clients can be jointly calculated on the server-side. Fig.~\ref{fig:Norm_Fed_Contrast} illustrates the mechanism of our correlation regularizer under the federated learning paradigms. 
Under the federated learning paradigm, in the client optimization, the local steps presented in Fig.~\ref{pipeline} enforce the instance embedding to be close to positive class embeddings. On the server side, the correlation regularizer makes the positive and negative class embeddings stay away from each other.
{In general, there tend to be more negative class pairs than positive class pairs. In specific tasks, such as the AmazonCat dataset containing 13000 labels, a label typically triggers a positive class pair with approximately five other labels on average. Moreover, each label can potentially form a negative
class pair with most of the other labels within 13000. Consequently, the quantity of negative class pairs is significantly higher. Storing the negative class pairs necessitates 4GB of storage, which constitutes a substantial resource overhead, whereas storing the positive class pairs only requires 800MB. The burden of storing negative class pairs may prove to be overly taxing for practical FL scenarios.}

\begin{figure}[t]
    \begin{center}
    \includegraphics[width=1\columnwidth]{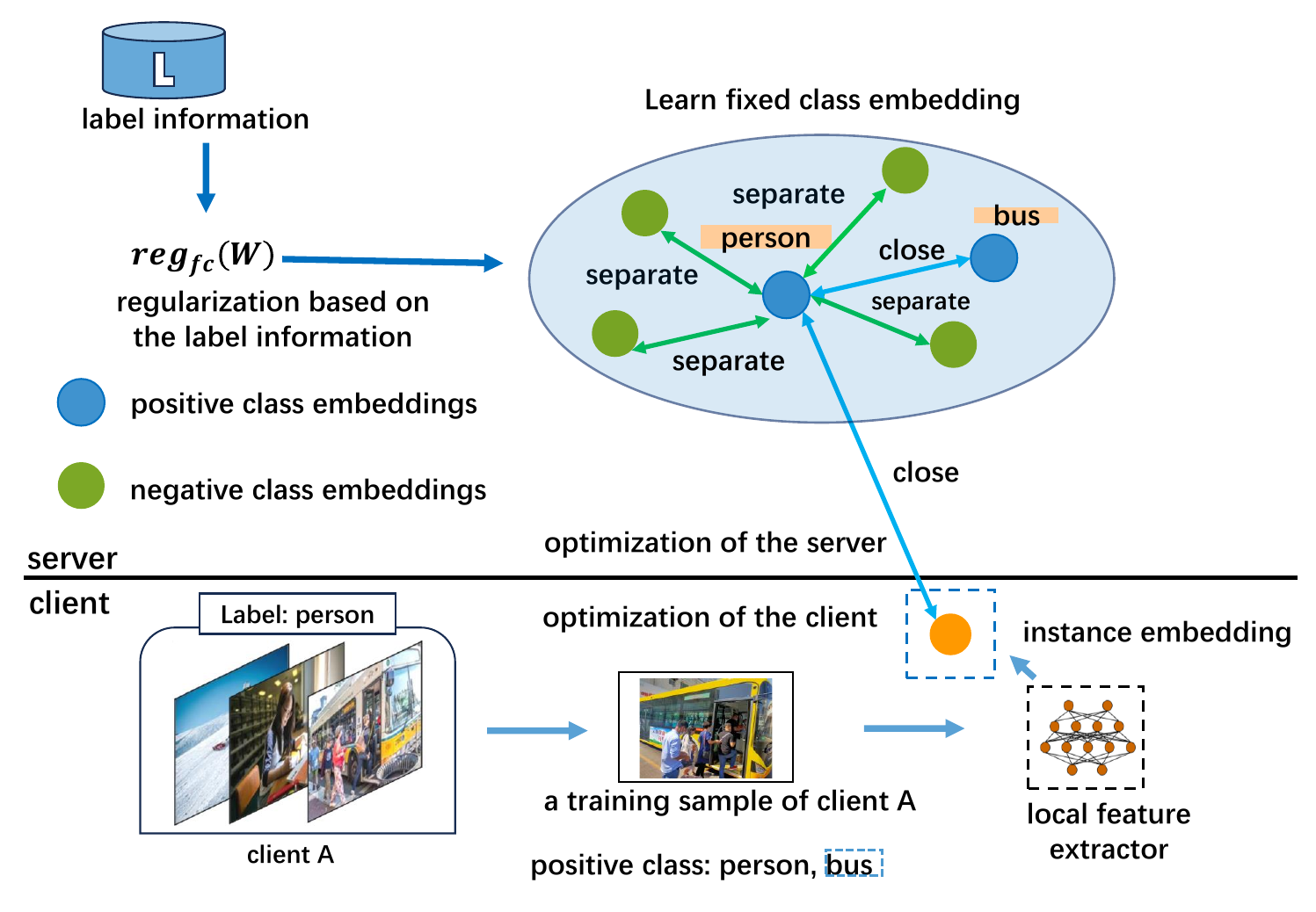}
    \end{center}
    \caption{{FedALC with a fixed class embedding matrix. On the server side, positive class embeddings are enforced to be close and separated from negative class embeddings. Then the fixed class embeddings are obtained; On the client side, the instance are enforced to approach its positive classes.}}
    \label{fixed_class_embedding}
\end{figure}

\subsection{Learning fixed class embedding (FedALC-fixed)}
In the above mentioned approach, the class embeddings are transmitted frequently between the server and clients. This may lead to the occurrence of privacy leakage, since the class embedding may contain the identity information of users.
To remedy this drawback and also reduce the communication overhead and computational cost, we propose a variant of our method to learn fixed class embeddings so that the transmission of class embeddings is performed only once. Since we want the class embedding matrix to be fixed and cannot be updated alternatively, the learned embedding matrix should be of high quality. This is achieved by using the positive class embedding to replace the corresponding instance embedding in both terms of Eq.~(\ref{client_loss_expected}) to learn a fixed class embedding matrix. According to the distribution of label sets, we define the correlation regularizer for learning the fixed class embedding matrix as follows:
\begin{equation}\label{fixed_reg}
\begin{split}
&\mathrm{reg}_{\mathrm{fc}}(W)
=\sum_{i \in[m]}\frac{|\mathcal{S}^{i}|}{|\mathcal{S}|}\left((\hat{\mathcal{R}}_{\mathrm{pos}}\left(\mathcal{S}^{i}\right)+\hat{\mathcal{R}}_{\mathrm{neg}}\left(\mathcal{S}^{i}\right)\right)\\
&= \frac{1}{n_{all}}\sum\limits_{i\in [m]}\sum\limits_{j\in [n_{i}]}\left(\ell^{\mathrm{pos}}_{cl}({L_{j}^{P}}, {L_{j}^{N}},W)+\ell^{\mathrm{neg}}_{cl}({L_{j}^{P}}, {L_{j}^{N}},W)\right)\\
&=\frac{1}{n_{all}} \sum\limits_{j\in [n_\mathrm{all}]} \left( \alpha \cdot\sum\limits_{y_j \in L_{j}^{P},\atop y_j^{\prime} \in L_{j}^{P} \And  y_j^{\prime}\neq y_{j}} \left(d(\mathbf{w}_{y_j}, \mathbf{w}_{y_j^{\prime}})\right)^{2} \right.\\ 
&\left.+\beta \cdot \sum\limits_{y_j \in L_{j}^{P},\atop y_j^{\prime} \in L_{j}^{N}}\left(\max \left\{0, \nu -d(\mathbf{w}_{y_j}, \mathbf{w}_{y_j^{\prime}})\right\}\right)^{2}\right).
\end{split}
\end{equation}
For each instance, the positive loss in Eq.~(\ref{fixed_reg}) encourages the positive embedding to be close and the negative loss encourages negative and positive embedding to be far away from each other. This process is illustrated in Figure \ref{fixed_class_embedding}, where the server learns a fixed class embedding matrix using regularizer Eq.~(\ref{fixed_reg}) and sends it to each clients. In the client optimization, the local step makes the instance embedding approach its positive class embeddings. The transmission of fixed class embeddings between clients and the server is only conducted once since the obtained class embeddings are unchanged during the process of federated learning.

\subsection{{Convergence and Optimality for Training}}
{Due to the $\max$ opreator, our loss function is Lipschitz continuity but non-smooth. According to \cite{conf/icml/HuangL0021}, under full clients participation setting and using FedAvg optimizer, FL converges to the global optimal solution at a linear rate with proper learning parameters.
That is, for $N$ clients satisfying $\epsilon$ precision (training loss function $L(w, x) \le \epsilon$), there are at least $T=\mathcal{O}  \left(\frac{N}{\lambda \eta_l \eta_g K}*\log{1/\epsilon} \right)$ communication rounds, where $\lambda$ is the minimal eignvalue of Gram matrix (aka pairwise data point gradient), $\eta_l$ and $\eta_g$ are learning rate at client and server, $K$ is the local interval. Since we also adopt the FedAvg opitmizer and the full participation setting, our proposed FedALC can also achieves a linear convergence rate. Our method focuses on spreading out the class embedding by exploiting label correlation, which avoids class embedding collapse and thus reduce the misclassification rate. FedALC doesn't consider new design for local optimizers or server-side model aggregation method. The enhancements of regularization on class embeddings are difficult to reflect in the convergence rate's order. }

\subsection{Label sets collection}\label{collect}
\begin{figure}[!t]
    \begin{center}
      \includegraphics[width=\linewidth]{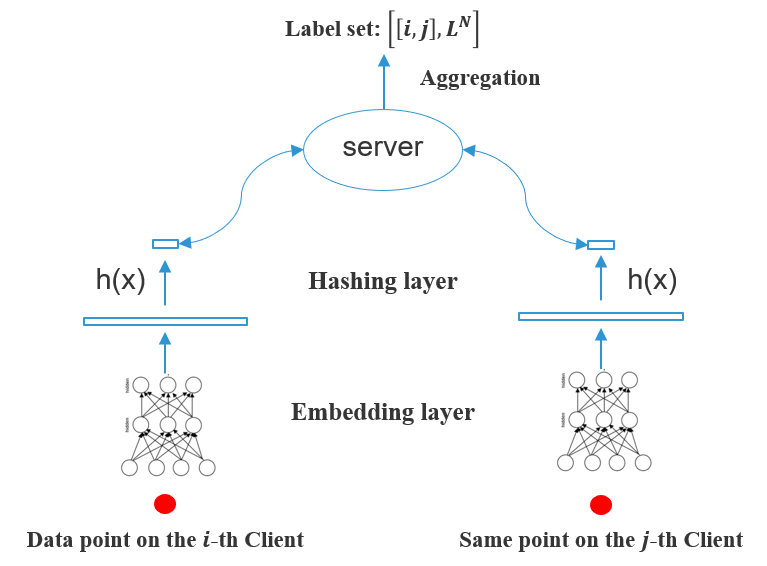}  
    \end{center}
    \caption{Label sets collection for each instance in the federated setting. Embeddings of the same data point on different clients are mapped as hash messages, which are sent to the server together with the label information. Then the server compare the messages and aggregate the labels that have the same message.}
    \label{collect label}
\end{figure}

There still remains a question on how to collect the labels for each instance. In multi-label classification, an instance may have multiple labels. Since each client is associated with a single label, the same instance may be distributed in different clients. To collect the labels for each instance, we propose to first use the irreversible hash algorithm, such as MD5~\cite{rivest1992md5} or SHA~\cite{wang2005finding} and all other Message Digest algorithms~\cite{bellovin2005deploying,coron2005merkle}, to map the instance embedding into the low-dimensional hash vector on the client side. Here, we assume that embeddings of the same instance on different clients are the same (i.e., the same feature extractor is adopted). Then the hash messages together with the associated labels are sent to the server. Finally, the received messages are compared and labels of the same messages are merged to obtain the label sets. The main procedure is illustrated in Figure~\ref{collect label}, where we transmit the hash messages instead of the original instance embeddings since: 1) instance embedding contains the user's identity information, and it is unsafe to exchange them between the server and clients. Some techniques (such as GAN\cite{9750393}) can be adopted to recover the user's data \cite{Sun_2021_CVPR}; 2) the embedding may be of large size and thus incur a large communication overhead.

{In this way, the server can indeed learn the distribution of labels of all data (in this paper, i.e., which client holds the data with some certain label) by collecting hash codes of data. However, the server cannot restore the original data samples from the hash codes due to the irreversibility of the hash function, and is not able to establish connections between data features and labels. Therefore, the model training needs to be done on each client, and the label distribution information obtained by the server plays the role of establishing the label correlation to avoid the model collapse.
In order to compute the correlation between labels, the hash codes of all the training data need to be transmitted from the clients to the server only once at the beginning of the FL process. The size of one hash code is small (32 Bytes using SHA-256 or 64 Bytes using SHA-512). Let's take VOC2012 dataset as an example, there are 11540 images and the client needs to transmit 0.15 MB hash codes to the server. Compared with the communication cost of the model parameter (23.5183M for ResNet-50), the cost incurred by hash codes is tolerable.}

{To mitigate the risk of label leakage, the client employs a hash function to encrypt the labels before uploading the encrypted instance and label to the server. This procedure ensures that the server solely accesses the correlation between labels, without any knowledge of their actual names, thereby rendering it impossible to establish the correspondence between instances and labels. This approach is particularly well-suited for real FL scenarios, safeguarding privacy and enhancing security.}

{The encryption and transmission process may lead to data leakage. All clients have the same initialization model and the same instance has the same hash code. Therefore, the malicious clients can infer other clients' data statistical from the embedding matrix $W$ under the setting of our method. To protect the privacy of the clients, we can conduct the embedding matrix $W$ training in a differential privacy manner. 
Specifically, differential privacy, a classical security method, is defined as follows:}
{
\begin{definition}(Defined in \cite{conf/tcc/DworkMNS06}.)
    A randomized algorithm $\mathcal{M}: \mathcal{D} \mapsto \mathcal{R}$ satisfies $\left(\epsilon, \delta\right)$-differential privacy (DP) if and only if for any two adjacent input datasets $D$ and $D^{'}$ that {differ} in a single entry and for any subsets of outputs $\mathcal{S} \subset \mathcal{R}$ it holds that
    \begin{equation}\nonumber
        Pr\left[\mathcal{M}: \mathcal{D} \in \mathcal{S} \right] \le e^{\epsilon} Pr\left[\mathcal{M}: \mathcal{D^{'}} \in \mathcal{S} \right] + \delta.
    \end{equation}
\end{definition}
}
{It has been demonstrated that with the DP-SGD\cite{conf/tcc/DworkMNS06} algorithm, the obatained embedding matrix $W$ is guaranteed to be $\left(\epsilon, \delta\right)-$differentially private\cite{conf/ccs/AbadiCGMMT016,journals/csur/LiuDSRFL21}, where $\epsilon$ and $\delta$ are the training privacy budget and training privacy tolerance hyperparameters, respectively. Following prior works in DP-SGD under the federated learning setting\cite{conf/iclr/McMahanRT018}, we can add Gaussian noise into the gradients during class embedding training at the server to provide a DP guarantee.}

{This work proposes two algorithms: FedALC and FedALC-fixed. FedALC is outlined in Fig \ref{pipeline}, Algorithm \ref{alg1} and elaborated upon in Section IV-B. The primary distinction between FedALC and FedALC-fixed lies in their optimization strategies for updating the class embedding matrix. Specifcally, FedALC employs one-step stochastic gradient descent to update the class embedding matrix within each round, whereas FedALC-fixed utilizes multiple steps of stochastic gradient descent to update the class embedding matrix before commencing the federated learning process, after which the class embedding matrix remains fixed throughout the subsequent training process. Fig \ref{fixed_class_embedding}, Section IV-C and Algorithm \ref{alg2} delineate the FedALC-fixed algorithm. The most prominent disparity between the FedALC and FedALC-fixed is highlighted in the “Training the fixed class embeddings ” section at the server in FedALC-fixed, indicated by blue coloring.} We separate the label collection part and the learning part in the light of clarity. Actually, the label collection can be performed together with the class embedding updates at the first round of communication.

\begin{algorithm}[tbp]
	\caption{Federated Averaging by exploring Label Correlations (FedALC) with collecting label sets} 
	\label{alg1} 
	\begin{algorithmic}[1]
		\STATE \textbf{Input:} $m$ clients, $\mathcal{S}^{i}$ at the $i$-th client.
		\STATE Server initializes parameters $\theta^{0}$, embeddings $W^{0}$, hashing function $H$.
		\STATE Server sends the initialization to clients.
		\STATE \textbf{Collecting label sets:}
		\STATE \textbf{for} i=1,2,3,...,m \textbf{do}:
		\STATE \quad The $i$-$th$ client downloads the initialization.
		\STATE \quad Compute hash vectors on the local data set:\\ \quad $M_{i}\gets H(g_{\theta}(\mathcal{S}^{i})) $
		\STATE \quad The $i$-th client sends the $M_{i}$  to the server.
		\STATE \textbf{end for}
	    \STATE Server integrates the $M_{i}$: $L \gets
	    \mathrm{merge}(M_{i})$
	    \STATE \textbf{Federated learning process:}
	    \STATE \textbf{for} t=0,1,2,...,T-1 \textbf{do}:
	    \STATE \quad Server sends $\theta^{t}$, $W^{t}$ to the $i$-th client
	    \STATE \quad \textbf{for} i=1,2,...,C \textbf{do}
	    \STATE \quad \quad $\left(\theta^{t}_{i}, \mathbf{w}^{t}_{y_{i}}\right) \leftarrow\left(\theta^{t}_{i}, \mathbf{w}^{t}_{y_{i}}\right)-\eta \nabla_{\left(\theta^{t}, \mathbf{w}^{t}_{y_{i}}\right)} \hat{\mathcal{R}}_{\text {pos }}\left(\mathcal{S}^{i}\right)$
	    \STATE \quad \quad $\text{where } \hat{\mathcal{R}}_{\text {pos }}\left(\mathcal{S}^{i}\right)=\frac{1}{n_{i}} \sum_{j \in\left[n_{i}\right]} \ell_{\mathrm{cl}}^{\mathrm{pos}}(\mathbf{x}^{i}_{j}, w^{t}_{y_i}).$
	    \STATE \quad \quad 
	    The $i$-th client sends $(\theta^{t}_{i}, \mathbf{w}^{t}_{y_{i}})$ to the server.
	    \STATE \quad \textbf{end for}
	    \STATE \quad Server updates the model parameters by averaging :\\
	    \quad $\theta ^{t+1}=\frac{1}{C} \sum\limits_{i\in [C] }\theta^{t}_{i}$,
	    \STATE \quad Server merges the class embedding \\ \quad and obtains {$\tilde{W}^{t+1}=W^{t}$}
	    \STATE \quad$ W^{t+1} \leftarrow \tilde{W}^{t+1}- \lambda \eta \nabla_{\tilde{W}^{t+1}} \operatorname{reg}_{\mathrm{c}}\left(\tilde{W}^{t+1}\right)$.
	    \STATE \textbf{end for}
	    \STATE \textbf{Output:} $\theta^{\mathrm{opt}},W^{\mathrm{opt}}$.
	\end{algorithmic} 
\end{algorithm}

\begin{algorithm}[tbp]
	\caption{Federated Averaging by exploring Label Correlations with fixed embeddings (FedALC-fixed)} 
	\label{alg2} 
	\begin{algorithmic}[1]
		\STATE \textbf{Input:} $m$ clients, $\mathcal{S}^{i}$ at the $i$-th client.
		\STATE Server initializes parameters $\theta^{0}$, embeddings $W^{0}$, hashing function $H$.
		\STATE Server sends the initialization to clients.
		\STATE \textbf{Collecting label sets:}
		\STATE \textbf{for} i=1,2,3,...,m \textbf{do}:
		\STATE \quad {The $i$-$th$ client downloads the initialization.}
		\STATE \quad Compute hash vectors on the local data set:\\ \quad $M_{i}\gets H(g_{\theta}(\mathcal{S}^{i})) $
		\STATE \quad The $i$-th client sends the $M_{i}$  to the server.
		\STATE \textbf{end for}
		\STATE Server integrates the $M_{i}$: $L \gets \mathrm{merge}(M_{i})$
		\STATE {\textbf{Training the fixed class embeddings:}}
		\STATE {\textbf{for} $t^{'}=0,1,2,...,T^{'}-1$ \textbf{do}}:
		\STATE {\quad$ \tilde{W}^{t^{'}+1} \leftarrow \tilde{W}^{t^{'}}- \lambda \eta \nabla_{\tilde{W}^{t^{'}}} \operatorname{reg}_{\mathrm{fc}}\left(\tilde{W}^{t^{'}}\right)$.}
		\STATE {\textbf{end for}}
		\STATE {The fixed class embedding matrix $W=\tilde{W}^{T^{'}}$.}
		\STATE \textbf{Federated learning process:}
		\STATE \textbf{for} $t=0,1,2,...,T-1$ \textbf{do}:
		\STATE \quad Server sends $\theta^{t}$, $W^{t}$ to the $i$-th client
		\STATE \quad \textbf{for} i=1,2,...,C \textbf{do}
		\STATE \quad \quad $\theta^{t+1}_{i}  \leftarrow \theta^{t}_{i} - \eta \nabla_{\theta^{t}} \hat{\mathcal{R}}_{\text {pos }}\left(\mathcal{S}^{i}\right)$
		\STATE \quad \quad $\text{where } \hat{\mathcal{R}}_{\text {pos }}\left(\mathcal{S}^{i}\right)=\frac{1}{n_{i}} \sum_{j \in\left[n_{i}\right]} \ell_{\mathrm{cl}}^{\mathrm{pos}}(\mathbf{x}^{i}_{j}, W).$
		\STATE \quad \quad 
		The $i$-th client sends $\theta^{t}_{i}$ to the server.
		\STATE \quad \textbf{end for}
		\STATE \quad Server updates the model parameters by averaging : $\theta ^{t+1}=\frac{1}{C} \sum\limits_{i\in [C] }\theta^{t}_{i}$.
		\STATE \textbf{end for}
		\STATE \textbf{Output:} $\theta^{\mathrm{opt}},W$.
	\end{algorithmic} 
\end{algorithm}

%% file: text/experiments.tex
\section{Experiments}
\label{sec:Experiments}

We empirically evaluate the proposed FedALC method on diverse multi-label datasets, including both the popular visual and some challenging textual datasets. We use cosine similarity to define the distance, i.e.,
\begin{equation}
{d}_{\cos }\left({x}, {y}\right)=1-{x}^{\top}{y}, \quad \forall{x},{y}\in \mathbb{R}^{d}.
\end{equation}
Given the specific distance function, the positive loss in the client can be given as follow:
\begin{equation}\label{postive loss}
    \ell_{cl}^{\mathrm{pos}}(\mathbf{x},y)={\max}({0, 0.9-g_{\theta}({\mathbf{x}})^{T} \mathbf{w}_{y}})^{2}.
\end{equation}
In our experiment, we follow the setting of positive loss in \cite{yu2020federated} that encourages all pairs of positive instance and its label $(x,y)$ to have dot product larger than $0.9$ in the embedding space.

\begin{table}[!t]
    \caption{A summary of the datasets used in multi-label image classification experiments.}
\centering
    \begin{tabular}{@{}lccccc@{}}    
        \toprule          
        {\scriptsize\textbf{Dataset}} &{\scriptsize\textbf{Labels}} &  {\scriptsize\textbf{Training images}} & {\scriptsize\textbf{Validation images}}& {\scriptsize\textbf{Test images}} \\ 
        \midrule
        COCO&80&73,873&8,208&4,0136\\
        
        VOC12&20&10,963&577&5,823\\
        
        VOC07&20&4,761&250&4,952\\

        {FLAIR}&{1,628}&{345,879}&{39,239}&{43,960}\\
        \bottomrule
    \end{tabular}
    \label{iamge_data_table}
\end{table}

\subsection{Datasets}
We use three benchmark datasets for multi-label image classiﬁcation, i.e., COCO\cite{lin2014microsoft}, VOC 2012\cite{everingham2010pascal}, VOC 2007\cite{everingham2010pascal} and {FLAIR\cite{conf/nips/SongGT22}} dataset. A summary of three datasets is presented in Table~\ref{iamge_data_table}.
The \textbf{COCO} dataset contains a total number of 122,218 images associated with 80 labels. We use the original training set and randomly select $10\%$ for validation. The original validation set is used for testing. The VOC 2012 (\textbf{VOC12}) dataset contains a total of 17,363 images associated with 20 labels. The original training set contains 11,540 images for training and validation, and we randomly select $5\%$ for validation and the rest for training. The original validation set containing 5,823 is used for testing. The VOC 2007 (\textbf{VOC07}) dataset contains a total number of 9,963 images associated with 20 labels. The original training set contains 5,011 images for training and validation. Similar to VOC12, we randomly select $5\%$ for validation and the rest for training. The original Validation set containing 4,952 is used for testing. {The \textbf{FLAIR} dataset contains a total of 429,078 images associated with 1,628 labels. The original training and validation set contain 385,118 images for training. The original test set containing 43,960 is used for testing. The original \textbf{FLAIR} dataset access are available at {https://github.com/apple/ml-flair} in \cite{conf/nips/SongGT22}.}

\begin{table}[!t]      
\caption{Summary of the textual datasets: N and M are the numbers of training and test instances, respectively. F is the feature dimension, L is the number of class labels, I/L is the average number of instances per label, and L/I is the average number of labels per instance.}
\centering
\label{tab1}
\setlength{\tabcolsep}{1mm}
    \begin{tabular}{@{}lcccccc@{}}    
        \toprule           
        {}&{N} & {M} & {F}& {L}& {I/L}& {L/I} \\ 
        \midrule
        EURLex&15539&3809&5000&3993&25.73&5.31\\
        
        Bibtex&4880&2515&1836&159&111.71&2.40\\

        {AmazonCat} &1,186,239&306,782&  203,882& 13,330&448.57&5.04\\
        
        {Amazon670K}&  485,176&150,875&   66,666&670,091&  5.11&5.99\\
        
        {WikiLSHTC} &1,778,351&587,084&1,617,899&325,056& 17.46&3.19\\
        
        \bottomrule
    \end{tabular}
\label{text_inf}
\end{table}
We also evaluate our algorithm on extreme multi-label text datasets: EURLex\cite{Menca2008EfficientPM}, Bibtex\cite{2008Multilabel}, {AmazonCat\cite{conf/bigcomp/GuanARL16}, WikiLSHTC\cite{journals/corr/PartalasKBAPGAA15} and Amazon670K\cite{conf/bigcomp/GuanARL16}}. 
These datasets are available in the Extreme Classification Repository\footnote{Most of the popular extreme classification datasets are available on the website: http://manikvarma.org/downloads/XC/XMLRepository.html}. A statistic of the sample, feature and class label of these multi-label textual datasets are shown in Table~\ref{text_inf}.

\subsection{Model architecture}
As mentioned earlier, we adopt the embedding-based classification model. For image datasets, we use \textbf{RESNET-50} \cite{he2016deep} as the backbone. The feature is then mapped into $512$-dimensional vector using linear transformation. Finally, the embedding is normalized so that its $l_2$-norm is $1$. In regard to the text datasets, high-dimensional sparse vector is embedded into $\mathbb{R}^{512}$ by embedding layer. The output is then passed through three linear layers with the size of $1024$, $1024$ and $512$. The first two layers use a ${ReLU}$ activation function. The output is then normalized to an unit vector. The instance embedding and class embedding are all $l_2$ normalized, and thus the predicted score belongs to the $[-1,1]$ interval.

\subsection{Implementation detail}
\begin{figure}[!t]
    \begin{center}
      \includegraphics[width=\linewidth]{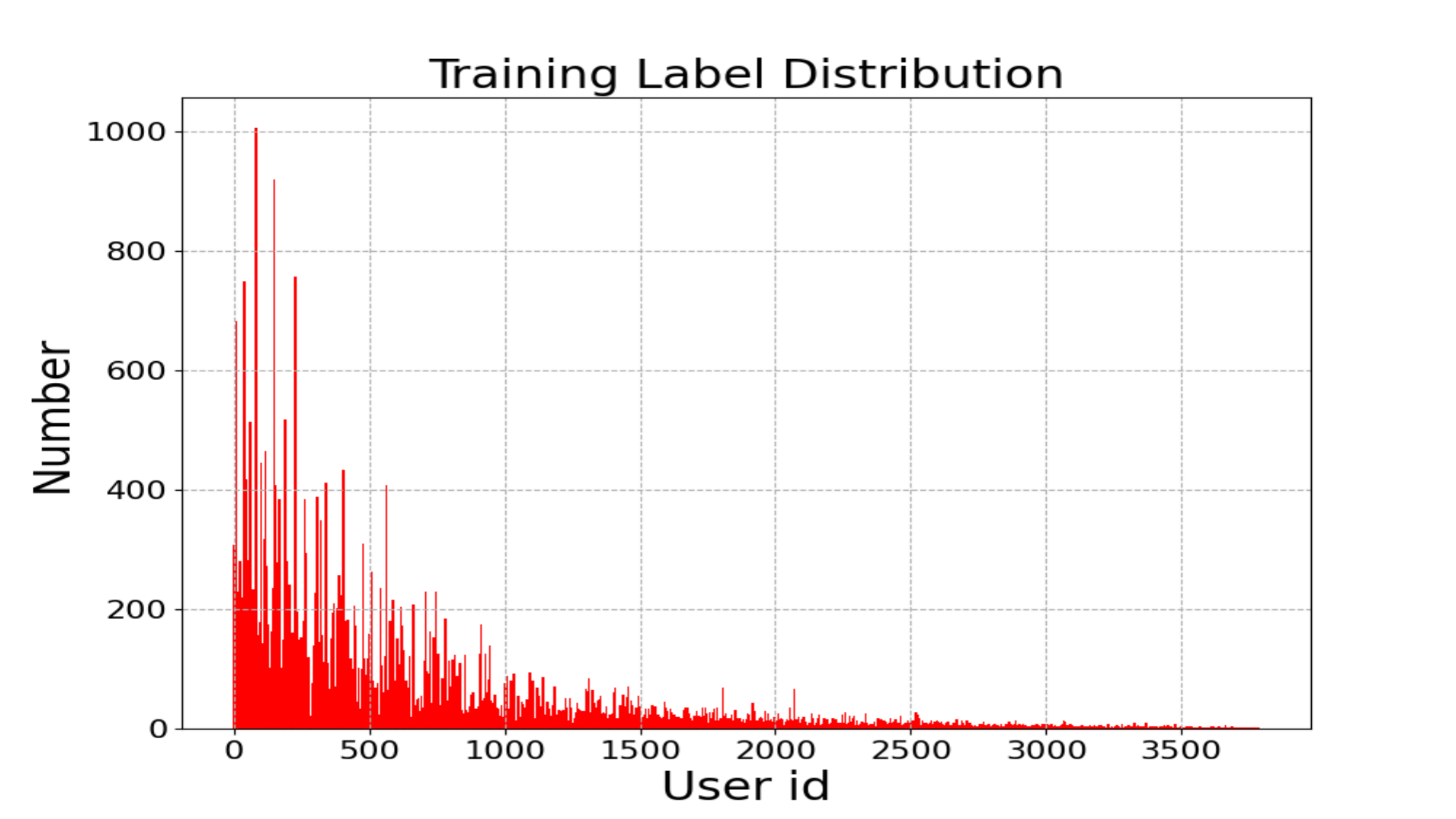} 
    \end{center}
    \caption{Visualization of number heterogeneity among users on EURlex dataset.}
    \label{fig:client_len}
\end{figure}

The methods are implemented using PyTorch, and all experiments are performed on server containing CPUs of 2.60GHz and GPUs of GeForce RTX 3090. SGD (stochastic gradient descent) is used to optimize the embedding-based model. 
{In this paper, each client corresponds to a label and the client only has the data whose label set includes the specified label. In this way, the number of clients equal to the number of labels. Since the data has multiple labels, the same data may appear in different clients.
From \cite{journals/corr/abs-1909-06335}, we can get that if Dirichlet distribution is used to split the training dataset over clients, the client would have data with the negative label.
In addition, Dirichlet distribution method assumes each class is independent, which conflicts with multi-label properties\cite{journals/corr/abs-2302-13571}.
Therefore, we donn't use Dirichlet distribution.}
We follow the setting in~\cite{yu2020federated} to use a large learning rate ($0.1$) when optimizing the instance embeddings on the clients, to accelerate the learning process. The learning rate on the server is set as $0.0001$. Most of the hyper-parameters in our FedALC method are the same as in FedAwS, where the balancing hyper-parameter $\lambda$ is tuned to achieve the best performance. Fig.~\ref{fig:client_len} is a statistic of the number of instances for all labels (clients) on the EURlex dataset, where we can see that the numbers vary significantly for different labels, and exhibit a long-tail distribution.
Such distribution also exists in other datasets in our experiments. In the following experiments, we let all clients participate in the communication to accelerate the training process. For image datasets, we run $100$ global communication rounds on COCO and {FLAIR} and $300$ rounds on VOC12 and VOC07. For text datasets, we set the global communication rounds to be $300$ and $500$ in the experiments of dynamic class embedding and fixed embedding variant, respectively (on both datasets).

\subsection{Compared approaches}
To show the effectiveness of the proposed method, we compare with the following approaches:
\begin{itemize}
\item{\textbf{SLEEC\cite{bhatia2015sparse}:}} Training by employing SLEEC in the oracle mode, which has access to both positive and negative embeddings. \item{\textbf{FedAvg\cite{mcmahan2017communication}:}} Training by employing FedAvg with only positive loss. In order to reduce the possibility of collapsing as much as possible, we initialize the class embedding matrix that is almost spreadout.
\item{\textbf{FedAwS}\cite{yu2020federated}:} Training by employing FedAwS with stochastic negative mining and the balancing hyper-parameter $\lambda$ controlling the effect of the spreadout on the class embedding matrix.
\item{\textbf{FedAvg-fixed:}} Training by employing FedAvg with only positive loss with the fixed class embedding, which is generated by random.
\item{{\textbf{FedMLH\cite{journals/corr/abs-2110-12292}}:}} {leverages label hashing to reduce the model size and achieve low communication cost. The class embedding matrix adopted are the same as FedAvg-fixed.}
\item{{\textbf{FLAG\cite{journals/corr/abs-2302-13571}}:}} {proposes a novel aggregation method based on label distribution and occurrence in the client dataset. The class embedding matrix adopted are the same as FedAvg-fixed.}
\end{itemize}

\subsection{Evaluation criteria}
For multi-label classification datasets, especially text datasets, even though the label space is very large, each instance only has several relevant labels. Hence, it is crucial to choose the short and most relevant label list for each test instance and evaluate the prediction quality, where the label list is the short ranked list of potentially relevant labels. The classic evaluation metric for the multi-label text classification is the top $K$ precision ($P@K$). We use this metric to evaluate the quality of the prediction with $K=1,3,5$. We denote $y\in \{ 0, 1 \}^{L}$ as the vector of true labels of an instance, and $\hat{y}\in R^{L}$ as the algorithm-predicted score for the instance. Then the metric can be defined as :
\begin{equation}
    P@K=\frac{1}{k} \sum \limits_{l\in r_{k}(\hat{y})} y_{l}
\end{equation}
where $r_{k}(\hat{y})$ is the set of rank indices of the truly relevant labels among the top-$k$ portion of the algorithm-predicted ranked list for an instance. In multi-label image classification, we use the classic mean average precision~(MAP)\cite{wang2020multi} and also $P@K$ as the evaluation criteria.

\begin{table}[!t]
\caption{MAP($\%$) and Top-k($\%$) on multi-label image datasets.}
\centering
\setlength{\tabcolsep}{5mm}
    \begin{tabular}{@{}l|ccc@{}}    
         \toprule       
        {Dataset}&{Method}&{MAP}&{P@3} \\ 
        \hline  
        \multirow{6}{*}{COCO} &FedAwS&30.33&\textbf{25.72}\\

        ~&{FLAG}&{26.55}&{19.60}\\

        ~&{FedMLH}&{26.14}&{20.98}\\
       
        ~&\textbf{FedALC~(Our)}&\textbf{30.35}&25.60\\
      
        \cline{2-4}
        ~&FedAvg-fixed&26.86&20.26\\
       
        ~&\textbf{FedALC-fixed~(Our)}&\textbf{30.22}&\textbf{25.03}\\
         \hline  
        \multirow{6}{*}{VOC12} &FedAwS&37.70&18.04\\

        ~&{FLAG}&{35.16}&{16.33}\\

        ~&{FedMLH}&{35.24}&{16.24}\\
     
        ~&\textbf{FedALC~(Our)}&\textbf{40.78}&\textbf{20.41}\\
       
         \cline{2-4}
   
        ~&FedAvg-fixed&37.19&17.40\\
       
        ~&\textbf{FedALC-fixed(Our)}&\textbf{43.33}&\textbf{19.77}\\
        \hline  
        \multirow{6}{*}{VOC07} &FedAwS&41.65&21.16\\

        ~&{FLAG}&{37.54}&{19.86}\\

        ~&{FedMLH}&{38.02}&{20.16}\\
  
        ~&\textbf{FedALC~(Our)}&\textbf{42.73}&\textbf{23.00}\\
   
        \cline{2-4}
     
        ~&FedAvg-fixed&41.54&21.56\\
      
        ~&\textbf{FedALC-fixed~(Our)}&\textbf{42.66}&\textbf{21.56}\\
        \hline  
        \multirow{6}{*}{{FLAIR}} &{FedAwS}&{16.78}&{5.15}\\

        ~&{FLAG}&{13.84}&{3.86}\\

        ~&{FedMLH}&{13.96}&{3.94}\\
  
        ~&\textbf{FedALC~(Our)}&\textbf{17.34}&\textbf{5.92}\\
   
        \cline{2-4}
     
        ~&{FedAvg-fixed}&{14.35}&{4.12}\\
      
        ~&\textbf{FedALC-fixed~(Our)}&\textbf{16.96}&\textbf{5.56}\\
        \bottomrule
    \end{tabular}
\label{result_image}
\end{table}

\subsection{Experiments on multi-label image datasets}
The results are shown in Table \ref{result_image}. 
{FedALC improves the mean average precision (MAP) $1.00\%$, $8.17\%$, $2.35\%$ and $3.34\%$ over the best baseline FedAwS on COCO, VOC12, VOC07 and FLAIR datasets.}
From the table, we can see that: 1) since the number of labels is not very large for these datasets, the superiority of FedAwS compared with FedAvg is limited. This may be because the numbers of labels are not very large in these datasets, and it is easy to initialize a dispersed class embedding matrix, which is similar to the one obtained by simple spreadout, as adopted in FedAwS. By exploring label correlations, our FedALC consistently outperforms FedAvg under different criteria; 2) our method consistently outperforms FedAwS in terms of MAP, and is superior to FedAwS in most cases at $P@3$. In particular, we achieve a significant $8.17\%$ relative improvement in terms of MAP on VOC12.
we also find that the performance on fixed class embeddings is better than the dynamic one in the experiment of VOC12. The reasons may be that: 1) the high-quality fixed class embedding which is spreadout enough and contains label correlations may achieve satisfactory performance; 2) the number of labels is limited in the multi-label images datasets, where the number of classes is often lower than 100 (compared with the thousands in the text datasets). In this case, it is easy to train the class embeddings to be high quality, and thus better performance are achieved. Regarding the dynamic embeddings, optimizations of class embeddings try to push some of them to be close on the client, and disperse some of them on the server. This contradictory operations may get stuck in training process.
{In addtion to FedAwS, the other three baselines (FedAvg, FedMLH, FLAG) are all relatively weak because they donn't consider the collapsing of class embedding matrix introduced by the lack of data with the negative label. Conversely, since FedAwS mitigates class embedding collapse by spreading out each class pair's geometric distances, similar performance to FedALC can be achieved. It demonstrates that the class embedding matrix collapsing is the bottleneck of Federated Learning with only positive labels and FedALC can achieve better performance than FedAwS is due to the exploration of labels correlations.}

{To discuss the communication cost of our method, we evaluate the total volume of data transmitted by both the server and clients during FL training. The findings for multi-label image classification tasks are succinctly presented in table \ref{tab-com}.}
    
    \begin{table}[!t]      
        \caption{The size of data transmitted by the server and the client for multi-label image datasets (MB).}
        \centering
        \label{tab-com}
        \setlength{\tabcolsep}{1mm}
            \begin{tabular}{@{}lcccccc@{}}    
                \toprule           
                {}     &\multicolumn{2}{c}{FedALC}    & \multicolumn{2}{c}{FedALC-fixed}     & \multicolumn{2}{c}{FedAWS} \\
                \midrule 
                {}     &{server} &{client}            & {server}      &{client}              & {server}     &{client}  \\
                \midrule
                {FLAIR}&2940   &2300.4                &  2306.4       &  2300                & 2940         &2300.4  \\
                
                {VOC12}&6924   &6901.2                &  6900.1       &  6900                & 6924         &6901.2  \\
        
                {VOC07}&6924   &6901.2                &  6900.1       &  6900                & 6924         &6901.2  \\
                
                {COCO} &2430   &2316                  &  2301.3       &  2300                & 2430         &2316    \\
                
                \bottomrule
            \end{tabular}
    \end{table}

    {For the server, our proposed FedALC-fixed demonstrates the minimal volume of data transmitted. This reduction in data volume stems from the fact that under FedALC-fixed, the server transmits the class embedding matrix only once during the FL training, whereas under both FedALC and FedAWS,
    the server transmits the class embedding matrix in each round. The most substantial reduction in data volume is observed in the FLAIR task. This is attributed to FLAIR having the largest number of labels(1628) and thus the class embedding matrix being of the largest size($1628 \times 512$). In comparison, the VOC07 and VOC12 tasks have the fewest labels (20), resulting in the smallest matrix size ($20 \times 512$), which minimally impacts communication costs. Consequently, the reduction in data size is less apparent for VOC07 and VOC12. In summary, when the size of the class embedding matrix is large, the proposed FedALC-fixed effectively mitigates communication cost for the server.}
    
    {For the client, FedALC-fixed also achieves minimal data volume in communication. This is because, in the settings of FedALC-fixed, the client only transmits the model without any class embedding during each FL round, unlike FedAWS and FedALC. FedALC reduces the data volume of communication for the client to a lesser extent compared with the server. This is because, under FedALC and FedAWS, the client only transmits the class embedding vector of the specified label, while the server must transmit the class embedding matrix for all labels. The most substantial reduction in data volume is observed in the COCO task, attributed to its larger class embedding dimension (2048) compared with the dimension of 512 for other tasks. Consequently, we can draw a conclusion that when the dimension of the class embedding vector is large, the proposed FedALC-fixed effectively reduces the communication cost for the client.}

\begin{table}[tbp]
\caption{Precision@1,3,5($\%$) with $K=10$ on EURlex.}
\centering
\setlength{\tabcolsep}{5mm}
    \begin{tabular}{@{}l|cccc@{}}  
        \toprule           
        {Method}&{$\lambda$}&{P@1}&{P@3}&{P@5} \\ 
        \hline
         {SLEEC~(Oracle)}&{}&{{63.40}}&{{50.35}}&{{41.28}}\\
         {FLAG}&{}&{28.50}&{20.66}&{17.87}\\
         {FedMLH}&{}&{29.01}&{21.74}&{17.64}\\
         {FedAvg}&{}&{29.61}&{21.85}&{17.95}\\
        \hline
        \multicolumn{5}{c}{$\mathbf{K=10}$}\\
        \hline
        \multirow{6}{*}{FedAwS}&{1}&{46.25}&{32.94}&{27.01}\\
       
         {}&{10}&{49.92}&{36.65}&{29.94}\\
       
         {}&{100}&{47.05}&{34.65}&{27.16}\\

         {}&{150}&{44.07}&{31.39}&{25.76}\\
       
         {}&{200}&{40.04}&{28.64}&{23.61}\\

         {}&{500}&{38.37}&{27.14}&{22.48}\\
         
         \hline
         \multirow{6}{*}{\textbf{FedALC~(Our)}}&{1}&{\textbf{54.37}}&{\textbf{38.68}}&{\textbf{31.99}}\\

         {}&{10}&{54.06}&{38.44}&{31.79}\\
       
         {}&{100}&{52.05}&{37.18}&{30.86}\\

         {}&{150}&{50.54}&{36.01}&{29.93}\\
       
         {}&{200}&{50.04}&{35.42}&{29.27}\\

         {}&{500}&{47.66}&{34.22}&{28.25}\\
         
         \bottomrule
    \end{tabular}
\label{Eurlex_result}
\end{table}

\begin{table}[!t]      
\caption{Precision@1,3,5($\%$) with $K=5$ on Bibtex.}
\centering
\setlength{\tabcolsep}{5mm}
    \begin{tabular}{@{}l|cccc@{}}    
        \toprule          
        {Method}&{$\lambda$}&{P@1}&{P@3}&{P@5} \\ 
        \hline
        {SLEEC~(Oracle)}&{}&{{65.08}}&{{39.64}}&{{28.87}}\\
        {FLAG}&{}&{11.43}&{9.08}&{7.36}\\
        {FedMLH}&{}&{11.14}&{8.91}&{7.47}\\
        {FedAvg}&{}&{11.6}&{9.11}&{7.84}\\
         \hline
         \multicolumn{5}{c}{$\mathbf{K=5}$}\\
        \hline
         \multirow{3}{*}{FedAwS}&{1}&{16.42}&{11.91}&{10.45}\\
       
         {}&{10}&{26.79}&{18.05}&{14.58}\\
      
         {}&{100}&{48.31}&{30.85}&{23.66}\\

         {}&{150}&{48.42}&{30.78}&{23.42}\\
      
         {}&{200}&{49.22}&{30.74}&{23.11}\\

         {}&{500}&{45.65}&{28.09}&{21.43}\\
    
         \hline
         
         \multirow{6}{*}{\textbf{FedALC~(Our)}}&{1}&{{58.72}}&{{35.83}}&{{26.62}}\\

         {}&{10}&{\textbf{59.67}}&{\textbf{36.04}}&{\textbf{27.18}}\\
      
         {}&{100}&{59.31}&{34.86}&{26.35}\\

         {}&{150}&{55.16}&{33.27}&{25.97}\\
      
         {}&{200}&{48.68}&{29.18}&{22.64}\\

         {}&{500}&{40.15}&{24.31}&{18.86}\\
         
         \bottomrule
    \end{tabular}
\label{Bibtex_result}
\end{table}

\begin{table}[!t]      
\caption{Precision@1,3,5($\%$) with $K=50$ on \\ AmazonCat, Amazon670K, WikiLSHTC.}
\centering
\setlength{\tabcolsep}{7.4mm}
    \begin{tabular}{@{}l|ccc@{}}    
        \toprule
        \multicolumn{4}{c}{AmazonCat}\\
        \hline
        {Method}&{P@1}&{P@3}&{P@5} \\ 
        \hline
        {SLEEC~(Oracle)}&{{90.50}}&{{76.31}}&{{61.50}}\\
        {FLAG}&{59.51}&{40.62}&{30.90}\\
        {FedMLH}&{57.83}&{39.39}&{28.24}\\
        {FedAvg}&{59.31}&{41.19}&{29.83}\\
        {FedAwS}&{85.40}&{65.60}&{48.12}\\
        \hline
        {\textbf{FedALC~(Our)}}&{\textbf{86.92}}&{\textbf{66.68}}&{\textbf{48.92}}\\
        \hline
        \multicolumn{4}{c}{Amazon670K}\\
        \hline
        {SLEEC~(Oracle)}&{{35.11}}&{{31.33}}&{{28.58}}\\
        {FLAG}&{2.98}&{1.81}&{1.41}\\
        {FedMLH}&{3.21}&{1.91}&{1.52}\\
        {FedAvg}&{3.10}&{2.04}&{1.68}\\
        {FedAwS}&{\textbf{31.61}}&{28.22}&{24.89}\\
        \hline
        {\textbf{FedALC~(Our)}}&{{31.22}}&{\textbf{28.43}}&{\textbf{25.30}}\\
        \hline
        \multicolumn{4}{c}{WikiLSHTC}\\
        \hline
        {SLEEC~(Oracle)}&{{54.81}}&{{33.41}}&{{23.92}}\\
        {FLAG}&{6.32}&{2.83}&{1.74}\\
        {FedMLH}&{6.22}&{2.69}&{1.91}\\
        {FedAvg}&{6.43}&{2.75}&{1.86}\\
        {FedAwS}&{36.28}&{21.87}&{14.80}\\
        \hline
        {\textbf{FedALC~(Our)}}&{\textbf{36.54}}&{\textbf{22.09}}&{\textbf{15.17}}\\
         \bottomrule
    \end{tabular}
\label{AmazonCat_result}
\end{table}

\begin{figure}[!t]
    \begin{center}
      \includegraphics[width=\linewidth]{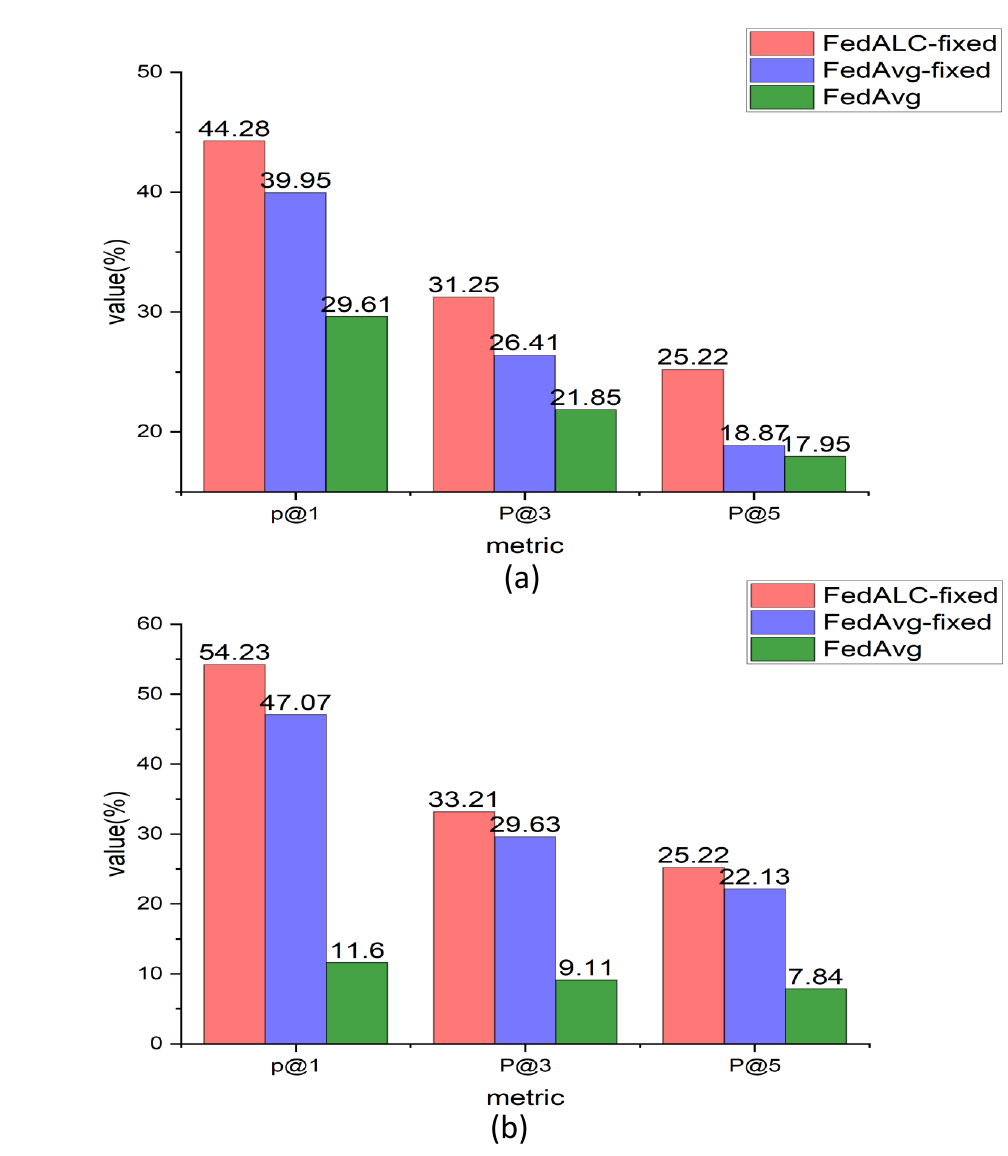}
    \end{center}
    \caption{Performance~($P@k$) comparison of FedALC-fixed and FedAvg-fixed, FedAvg on the EURlex and Bibtex datasets.}
    \label{fig:EURlex_Bibtex_fixed_C}
\end{figure}

\subsection{Experiments on extreme multi-label text datasets}
\begin{figure}[!t]
    \begin{center}
      \includegraphics[width=\linewidth]{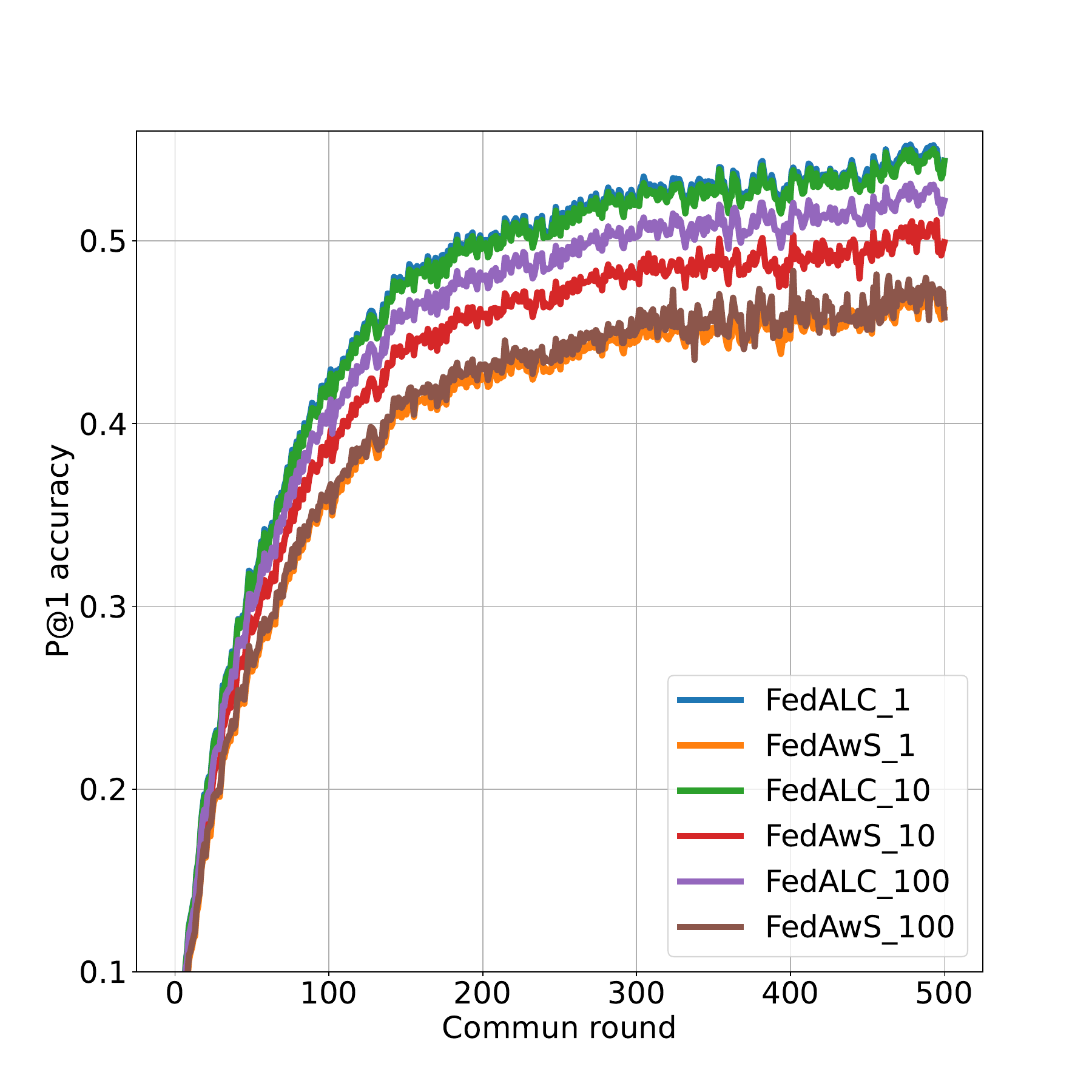} 
    \end{center}
    \caption{A comparison of our FedALC and the FedAwS counterpart under different hyper-parameter settings for $\lambda$ (e.g., ``\_100'' indicates $\lambda=100$) in terms of P@1($\%$) with $K=10$ on the EURlex dataset.}
    \label{fig:P@1 on Eurlex}
\end{figure}

\begin{figure*}[!t]
    \centering
    \subfloat[Precision@1,3,5 ($\%$) of FedALC on EURlex under ($\lambda=1,10,100, 150, 200, 500$) settings.]{
      \begin{minipage}{0.999\linewidth}
        \centering
        \includegraphics[width=0.3\linewidth]{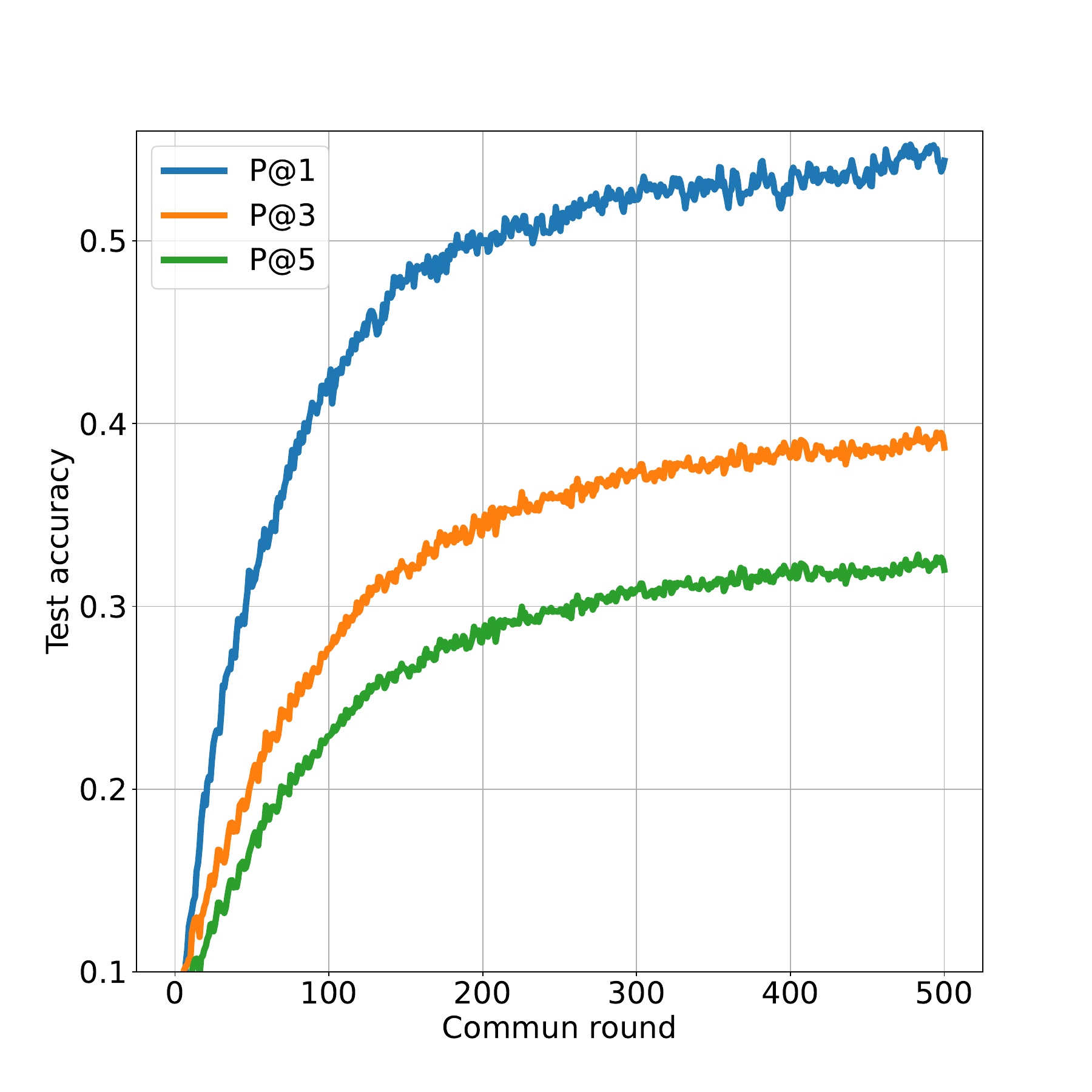}
        \includegraphics[width=0.3\linewidth]{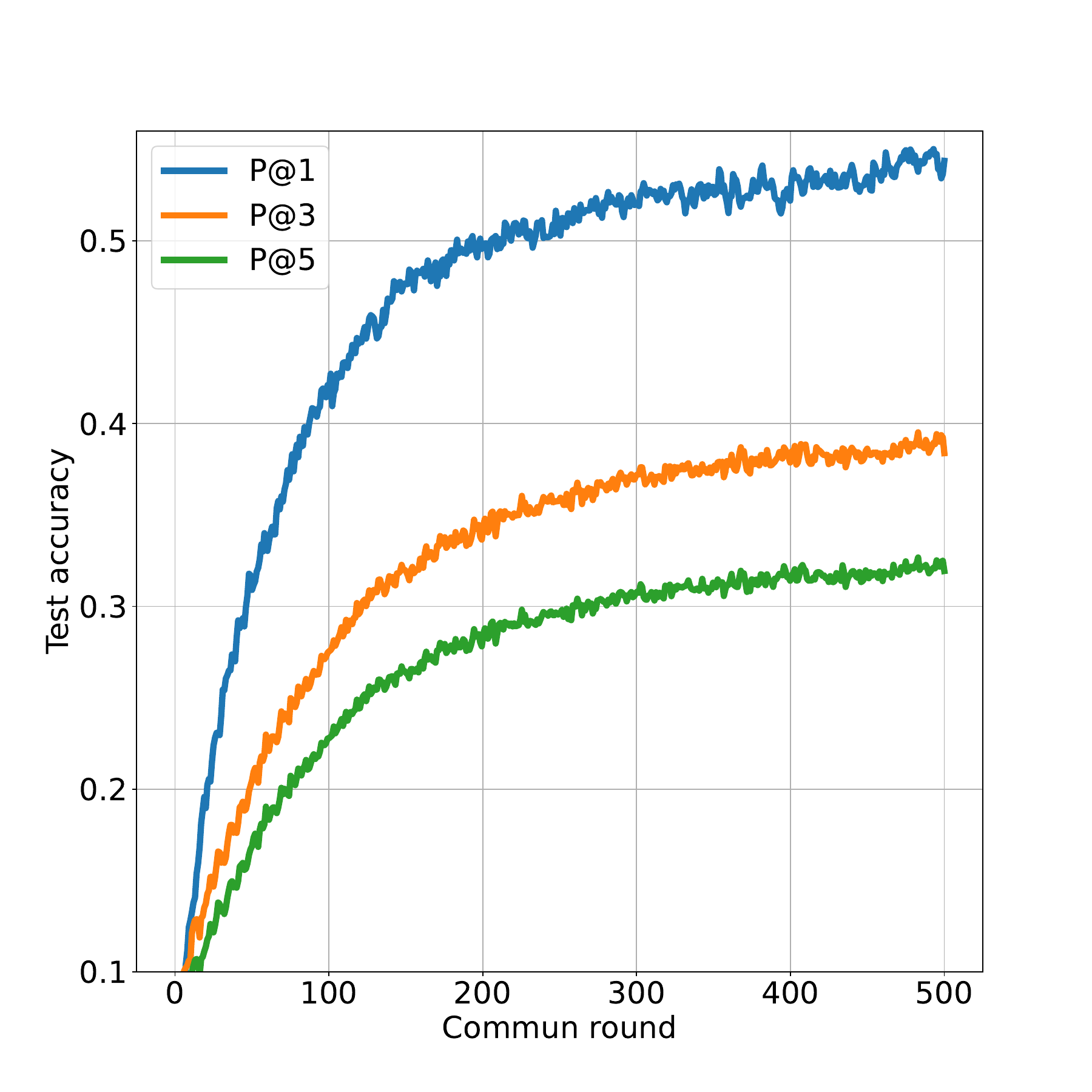}
        \includegraphics[width=0.3\linewidth]{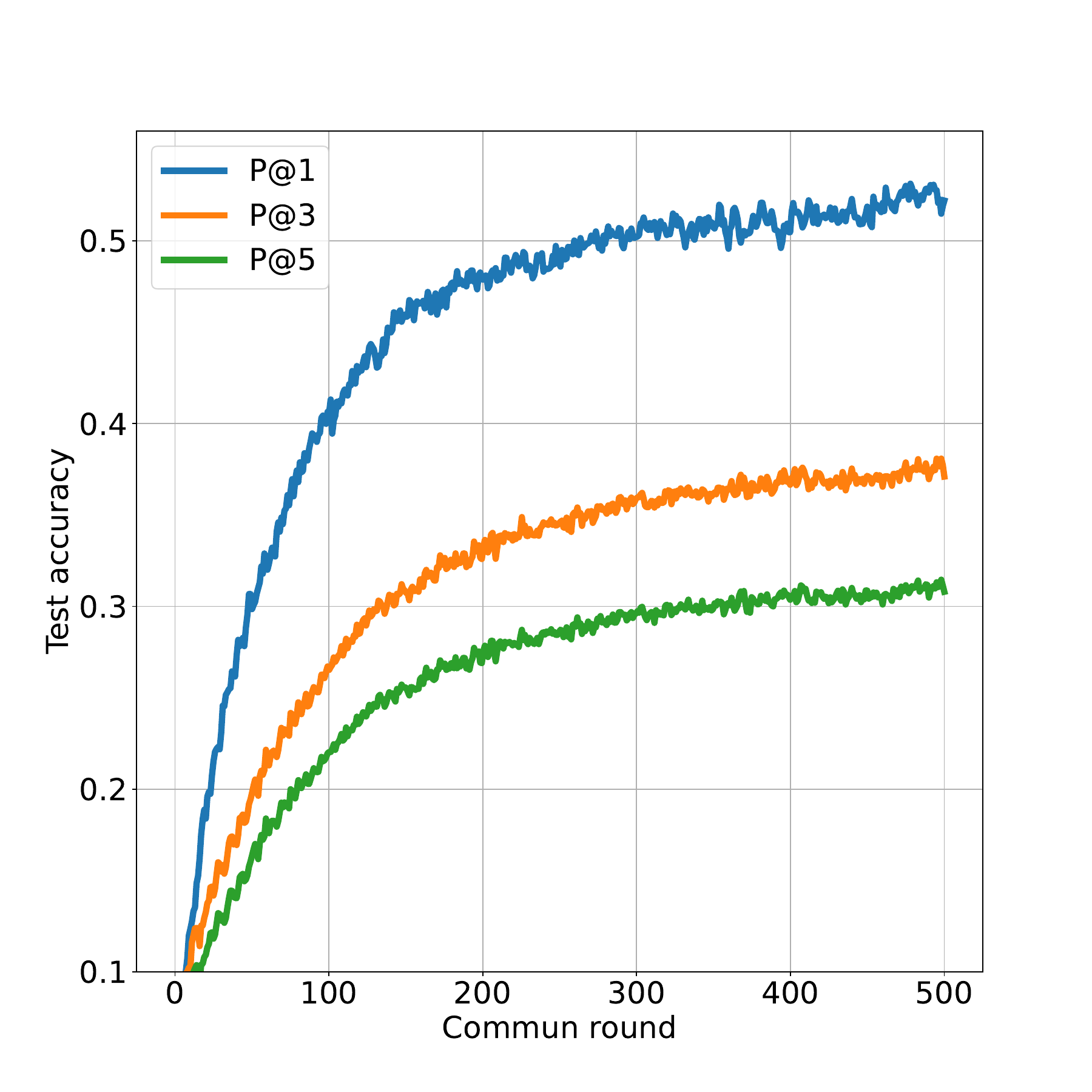}
        \includegraphics[width=0.3\linewidth]{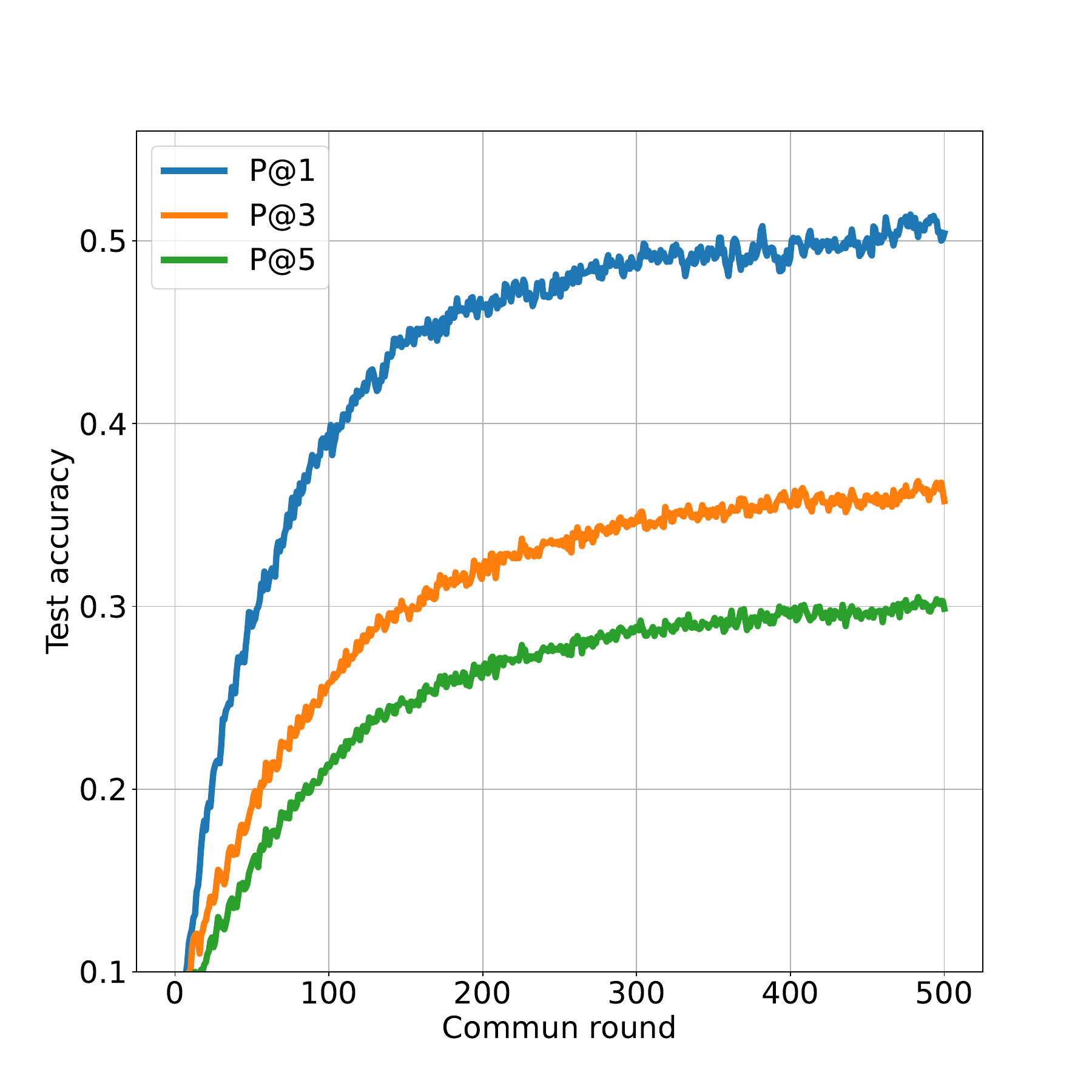}
        \includegraphics[width=0.3\linewidth]{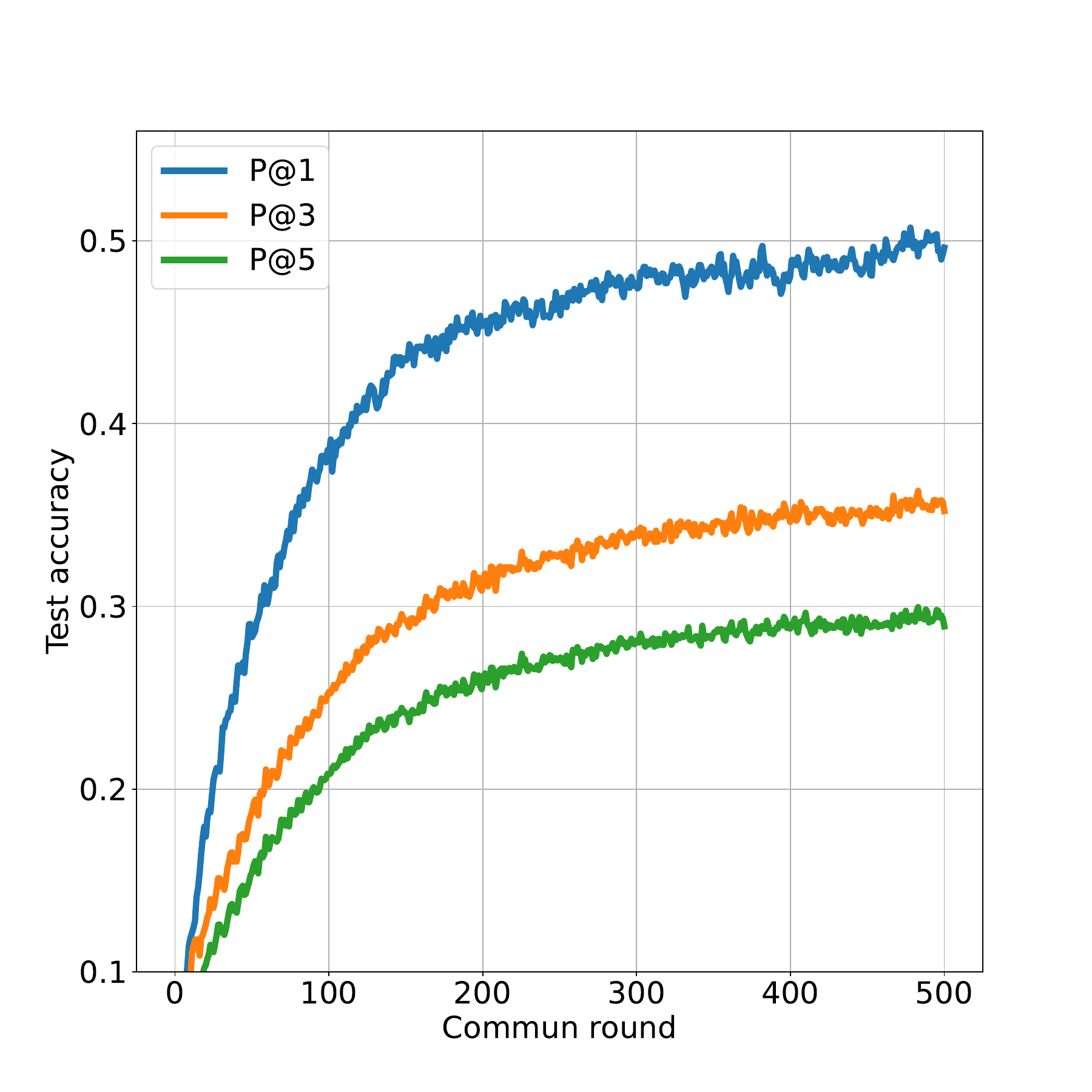}
        \includegraphics[width=0.3\linewidth]{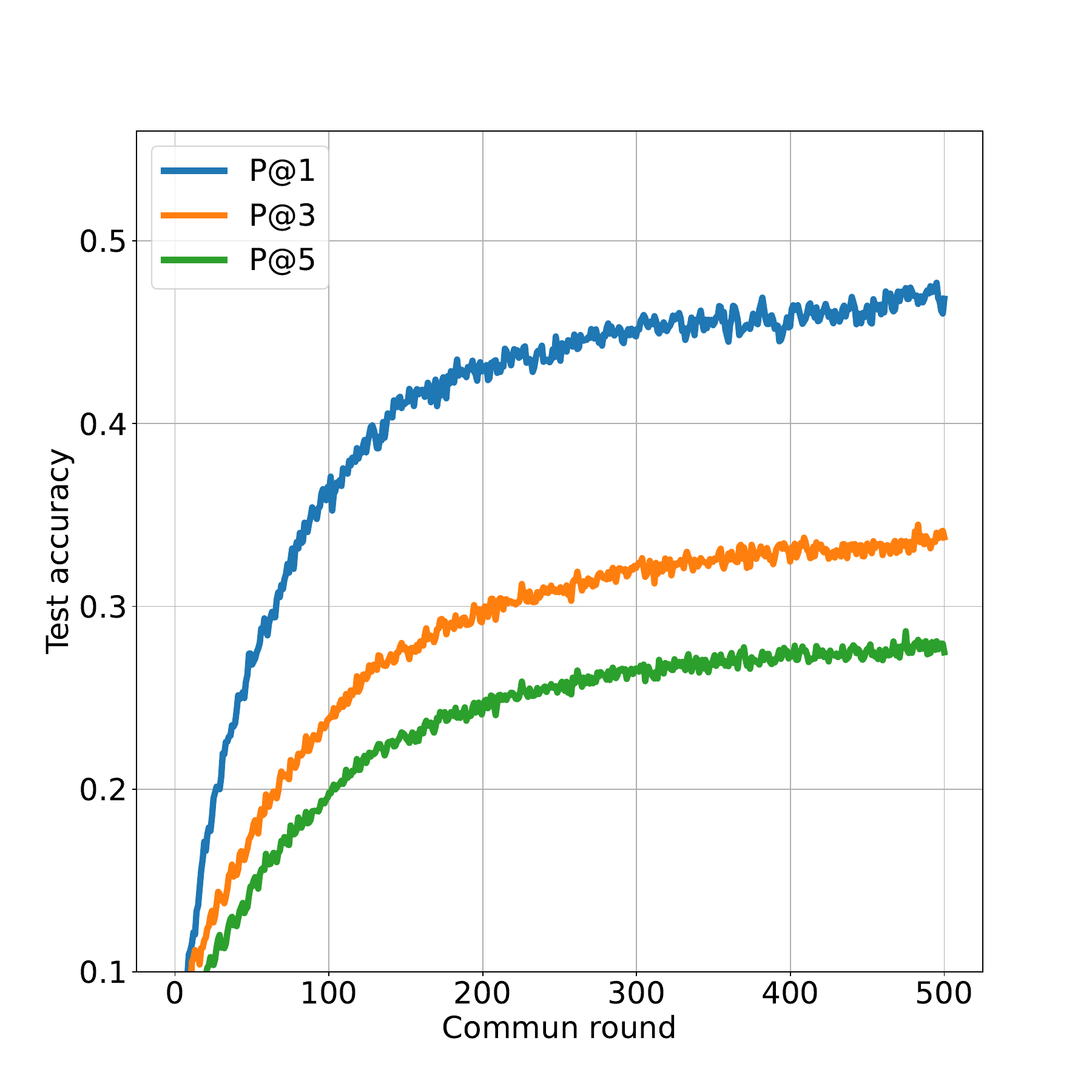}
        \label{FedALC-acc}
      \end{minipage}
    }
    \newline
    \subfloat[Precision@1,3,5 ($\%$) of FedAwS on EurLex under ($\lambda=1, 10, 100, 150, 200, 500$) settings.]{
      \begin{minipage}{0.999\linewidth}
        \centering
        \includegraphics[width=0.3\linewidth]{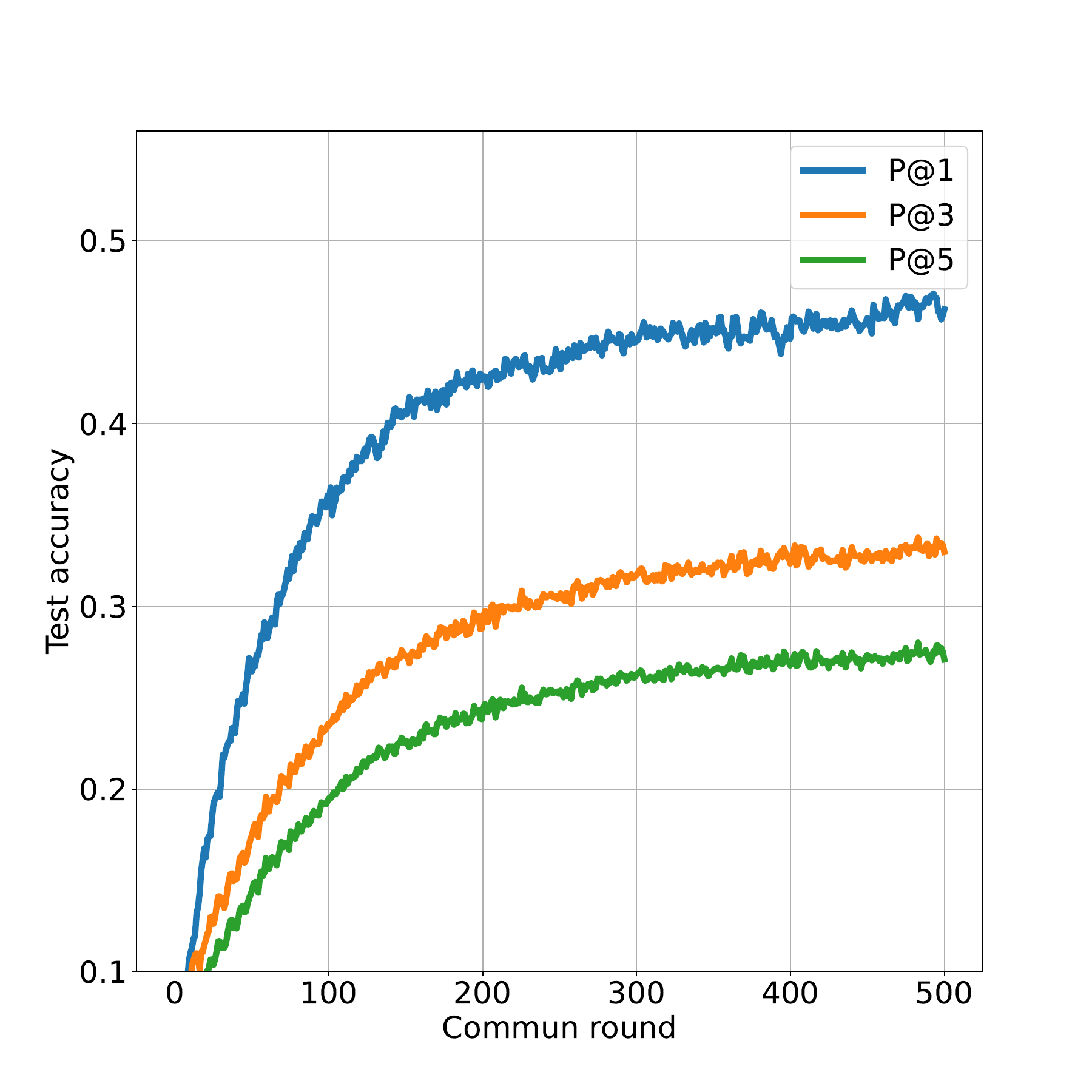}
        \includegraphics[width=0.3\linewidth]{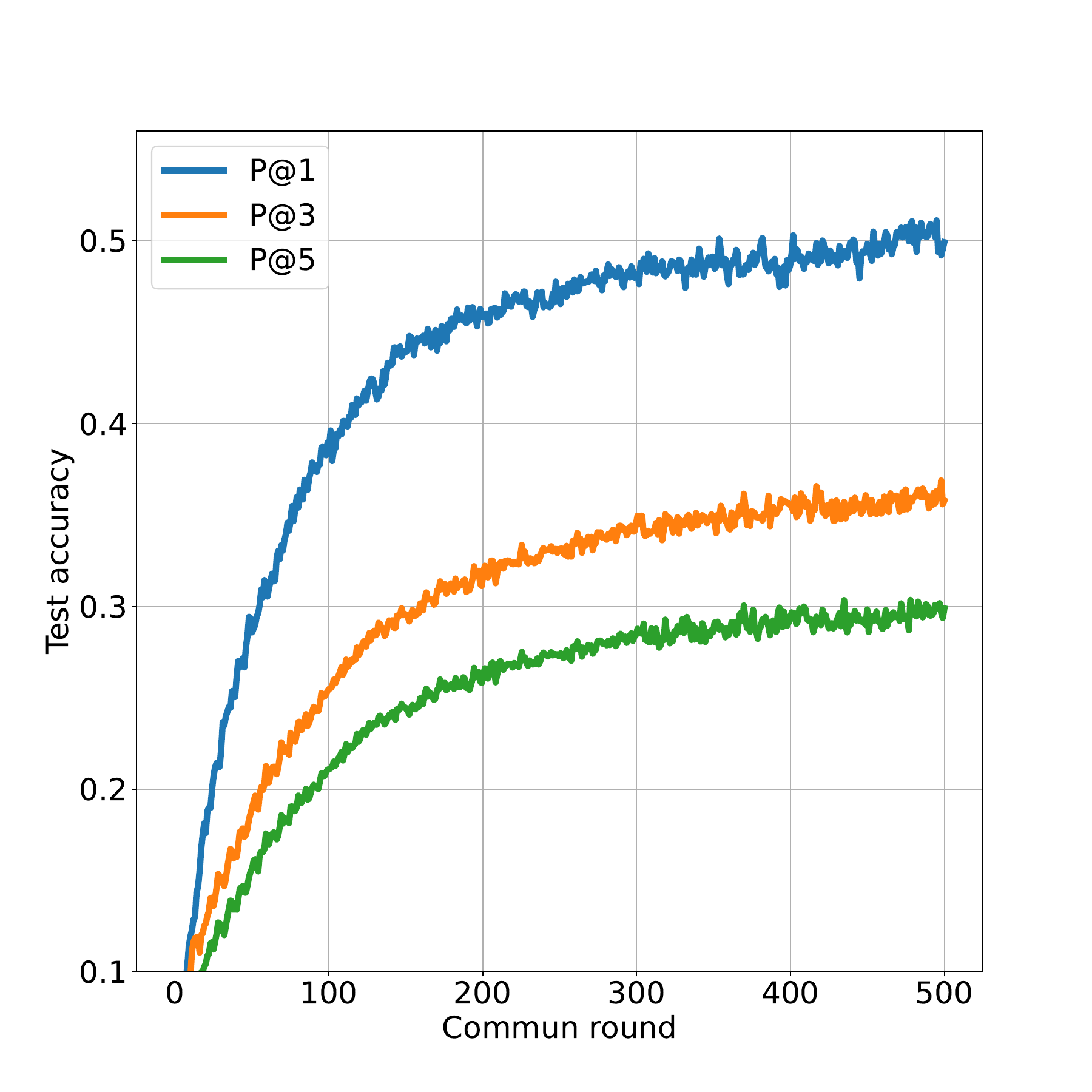}
        \includegraphics[width=0.3\linewidth]{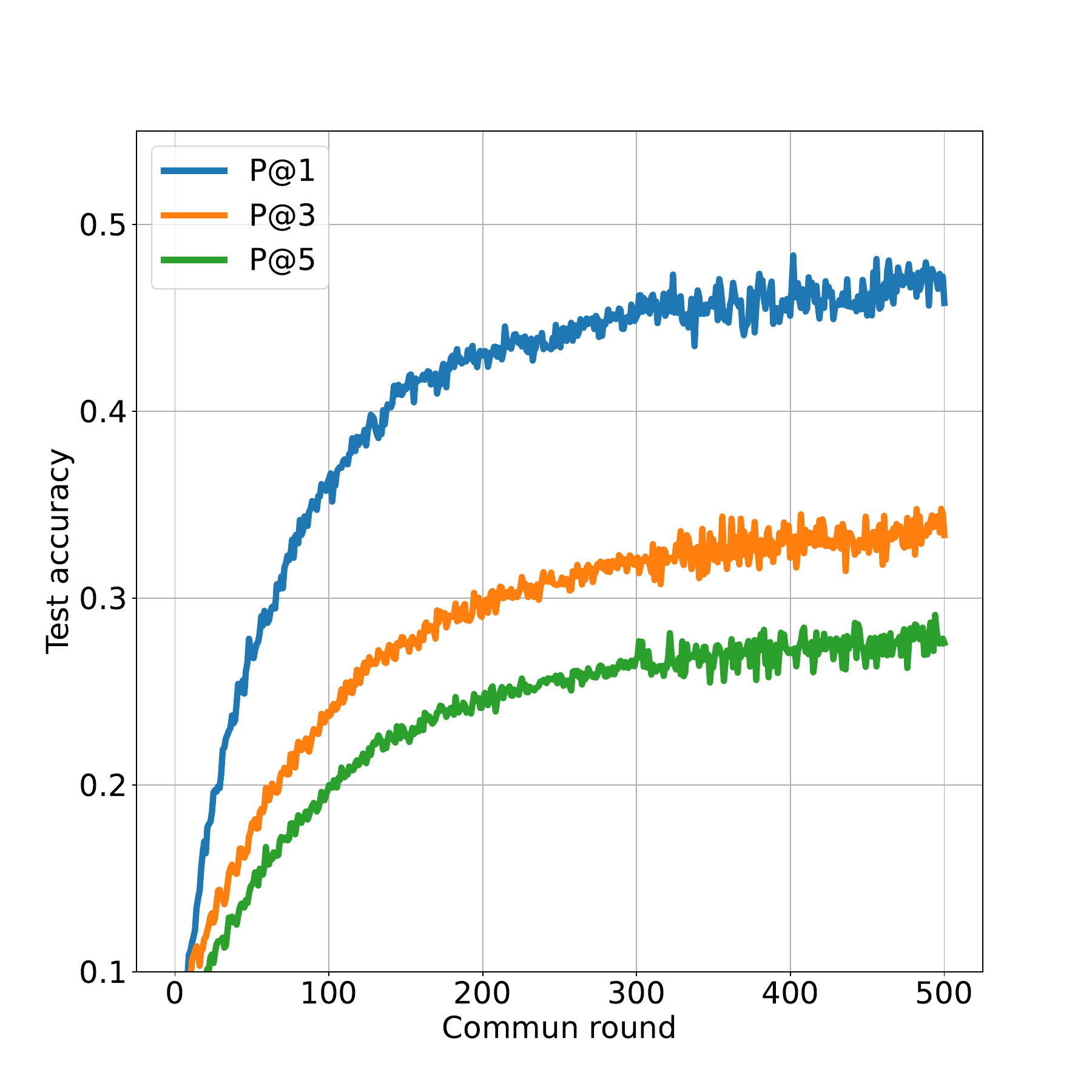}
        \includegraphics[width=0.3\linewidth]{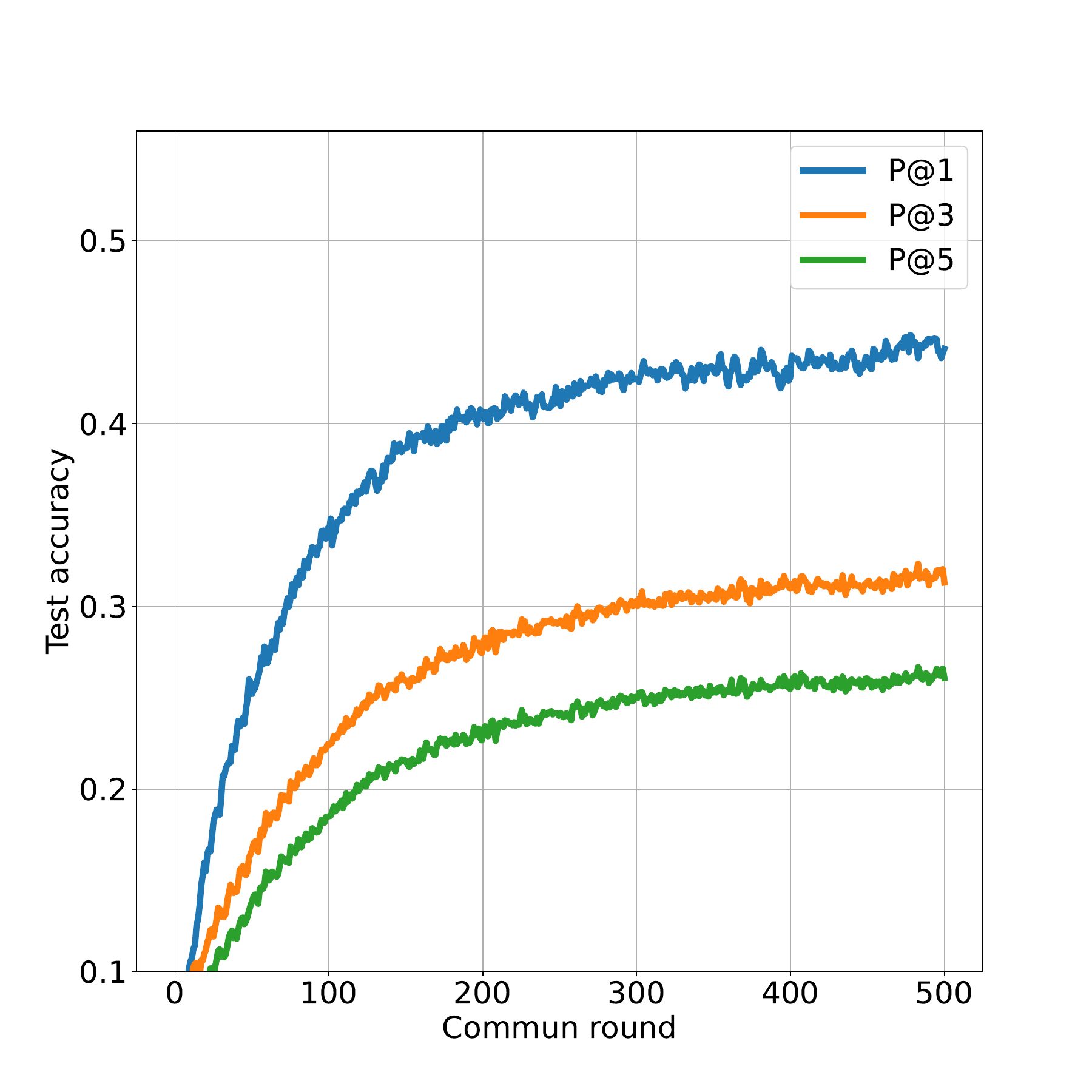}
        \includegraphics[width=0.3\linewidth]{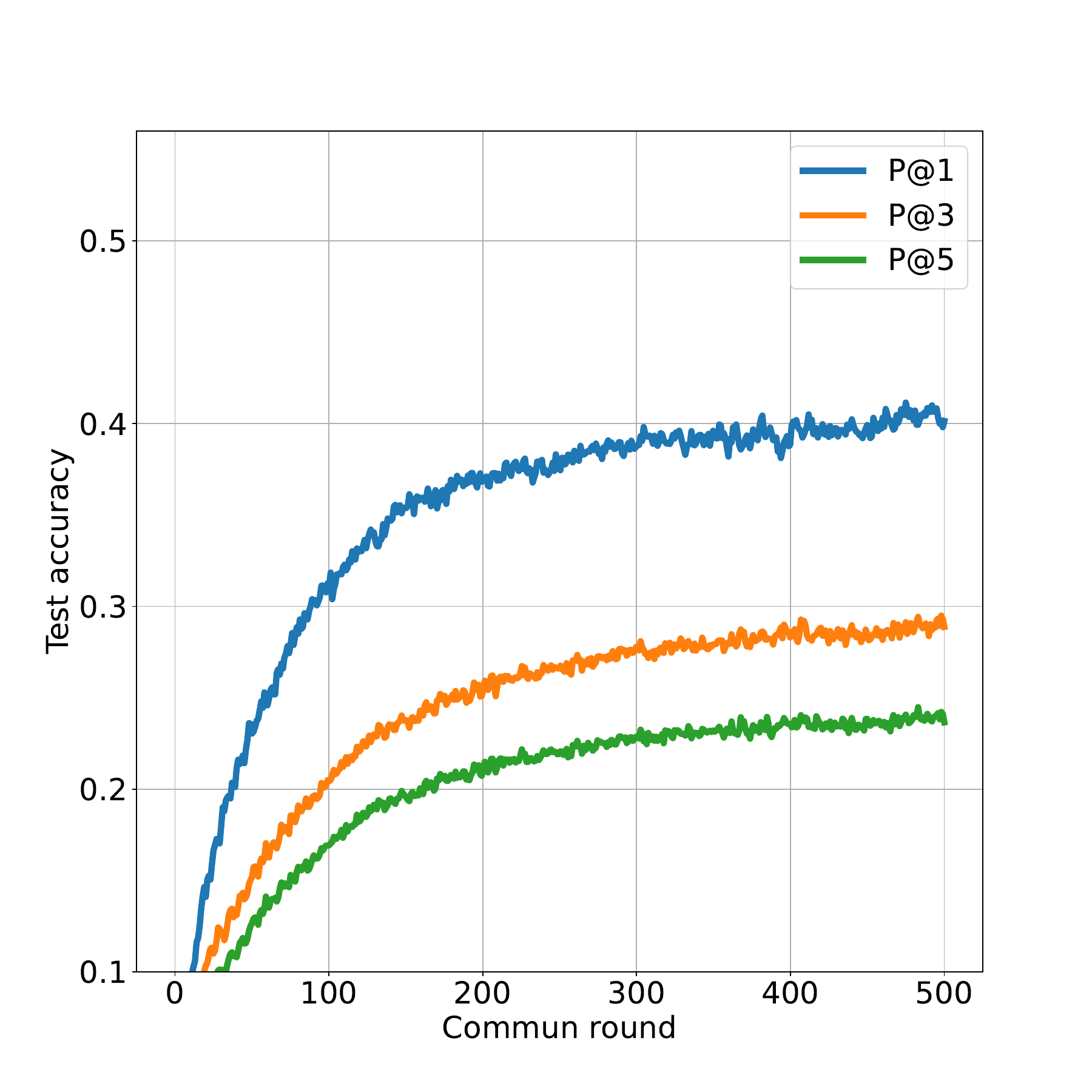}
        \includegraphics[width=0.3\linewidth]{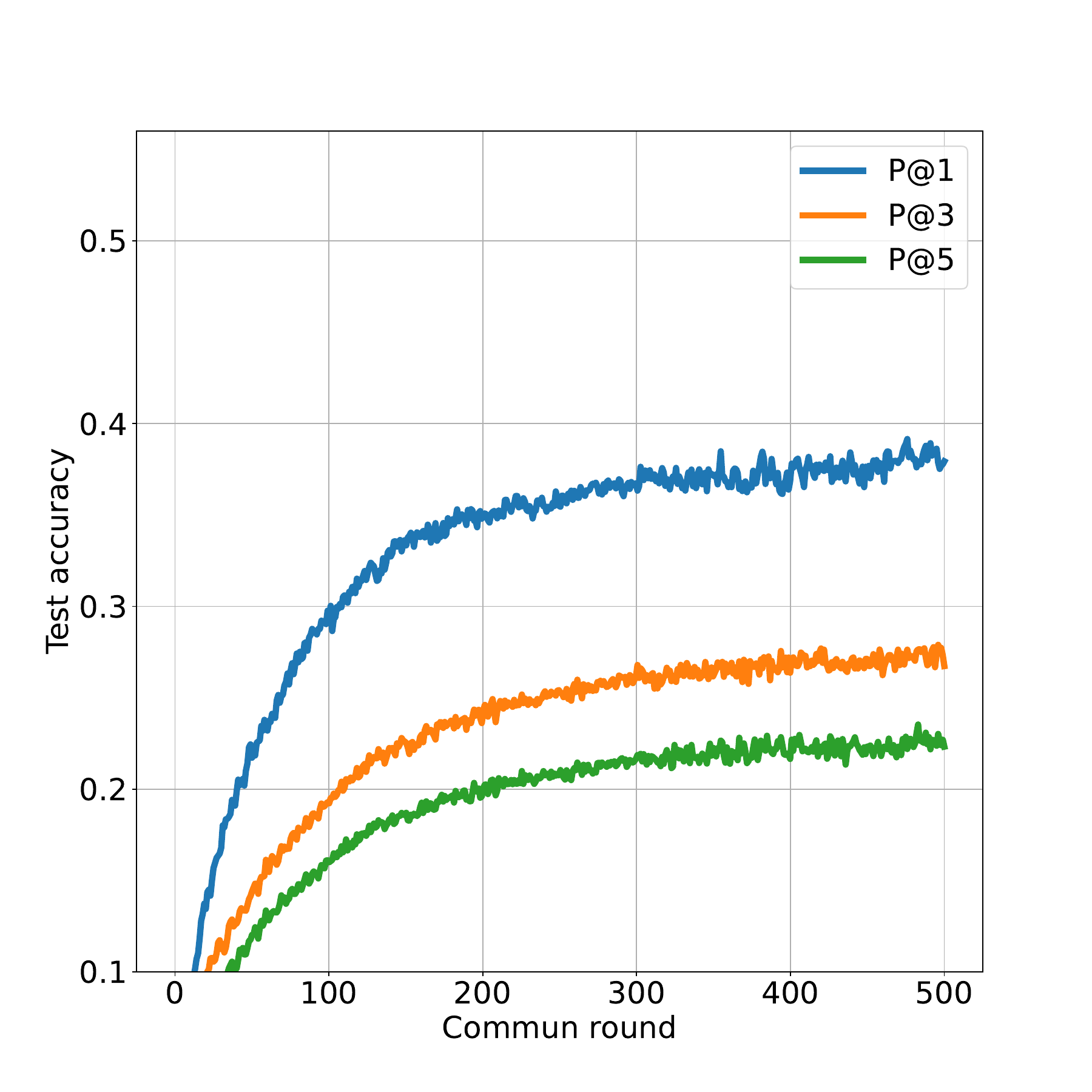}
        \label{FedAwS-acc}
      \end{minipage}
    }
    \caption{\small A comparison of our FedALC and the FedAwS counterpart under different hyper-parameters $\lambda$ in terms of Precision@1,3,5(\%) with $K = 10$ on the EURlex dataset.}
    \label{K-acc}
\end{figure*}

The overall results on EURlex are reported in Table \ref{Eurlex_result} and Fig.~\ref{fig:EURlex_Bibtex_fixed_C}(a), {the results on AmazonCat, WikiLSHTC and Amazon670K are summarized in Table \ref{AmazonCat_result}}, and the results on Bibtex are shown in Table \ref{Bibtex_result} and Fig.~\ref{fig:EURlex_Bibtex_fixed_C}(b). In particular, the learning curves of our FedALC and the most competitive FedAwS under different $\lambda$ in terms of $P@1$ on the Eurlex dataset are shown in Fig.~\ref{fig:P@1 on Eurlex}, and the corresponding results in terms of precision at different $k$ are shown in Fig.~\ref{K-acc}.

From the results of Bibtex and EurLex, we can see that: 1) {FedALC achieves higher precision $P@1,P@3, P@5$ than FedAwS (except for $\lambda=200$), the best test accuracy of FedALC is achieved with $\lambda=1$ and $\lambda=10$ on EURlex and Bibtex and} the {FedAvg, FLAG and FedMLH} algorithm prematurely converged and lead to relatively low accuracy because of the collapsing issue; 2) the proposed FedALC is much better than FedAvg in learning either dynamic or fixed class embeddings since our correlations regularizer can appropriately deal with the issue of collapsing; 3) {compared with FedALC, FedAwS is more sensitive to the hyper-parameter $\lambda$.} When the value of $\lambda$ is too small, the class embeddings cannot be well separated and this leads to collapsing issue and thus low accuracy. When $\lambda$ is too large, the class embedding matrix is over spreadout. The instance embedding is hard to approach corresponding positive class embeddings, and hence the performance may also be poor; 4) our FedALC can balance the dispersion and closeness of positive embeddings to some degree by exploring label correlations and automatically setting the weights for different label pairs. In particular, we achieve a significant $19.3\%$ relative improvement in terms of $P@1$ on the Bibtex dataset. FedALC disperses the matrix by taking label correlations into consideration, highly-correlated classes would be closer than those in FedAwS, and thus instance embedding would be easier to get close to its positive class embeddings and results in larger similarity. In regard to FedAwS utilizing stochastic negative mining, the given label and most of the selected neighbors are correlated with each other. Simply making them be arbitrary spreadout is thus sub-optimal. Whereas in our FedALC, more attention is paid on dispersing the embeddings of uncorrelated labels, and hence better performance can be obtained; 
{ 5) compared with multi-label image datasets, on EurLex and Bibtex datasets the improvement of FedALC over FedAwS are much stronger. The reason behind that is in these multi-label textual task, there are many labels are highly correlated\cite{jain2016extreme} and our method explores these correlations.}
This further verifies the benefit of exploring the label correlations.

{From Table \ref{AmazonCat_result}, for AmazonCat, Amazon670K and WikiLSHTC datasets, the situation is more complex. The baselines without considering class embedding matrix collapsing are relatively weak on all the three datasets. On AmazonCat dataset, FedALC improves $P@1$ ($P@3$, $P@5$) precision $1.78\%$ ($1.92\%$, $1.66\%$) over the best baseline FedAwS. On WikiLSHTC dataset, the improvements are relatively small ($0.72\%$, $1.01\%$, $1.07\%$ on $P@1$, $P@3$ and $P@5$ precision). On Amazon670K dataset, FedALC achieves a similar performance with FedAwS and on $P@1$ precision, the FedALC is lower than FedAwS. The reasons behind that are label correlations. For multi-label tasks, more instances per label means that label is positive correlated with more other labels. From Table \ref{text_inf}, we can observe that labels correlations on AmazonCat are strongest (the largest average number of instances per label) and the labels correlations on Amazon670K are the weakest (the least average instances per label). Our method explores the label correlation and achieves the largest improvement on the dataset with the strongest label correlations.} 

{From all the above results, we can find FedAvg, FLAG and FedMLH are relatively weak because that they donn't consider the class embedding collapsing introduced by the lack of negative label data. FedALC and FedAwS achieve relatively strong performance due to the fact that spreading out the class embedding matrix can effectly mitigate the class embedding matrix collapsing. FedALC achieves the highest accuracy in most cases, because we spreadout the class embedding with the consideration of label correlations.}

%% file: text/conclusion.tex
\section{Conclusion and Discussion}
\label{sec:Conclusion}

In this paper, we investigate the problem of the multi-label classification in the federated learning setting, where each client has only access to the local dataset associated with positive data, and propose a novel method that explores the label correlations to improve the quality of the class embedding matrix. Existing algorithm treats different labels equally, while our FedALC can explore label correlations by making the best use of label distributions. To improve safety and reduce the computational cost and communication overhead, we further propose a variant of our FedALC by learning a fixed class embedding matrix, where the server and clients only need to exchange class embeddings once. Moreover, a sophisticated strategy is designed for label collection. The experiments show that our algorithm is more adaptive and effective for the challenging federated multi-label setting.

In regard to limitations, simply using positive class embedding to replace the corresponding instance embedding may lead to information loss, especially in the variant of learning fixed class embedding. Exploring how to find better surrogate for instance embedding may further improve the performance and also accelerate the learning process.
Besides, we intend to verify the effectiveness of our method on more massive-scale extreme multi-label datasets.